\documentclass{article}
\IfFileExists{neurips_2026.sty}{\usepackage[preprint]{neurips_2026}}{\usepackage[margin=1in]{geometry}}
\usepackage[T1]{fontenc}
\usepackage[utf8]{inputenc}
\usepackage{amsmath,amsfonts,amssymb,amsthm,graphicx,booktabs,array,tabularx,xltabular}
\usepackage{microtype,xurl}
\usepackage{natbib}
\setcitestyle{numbers,square,comma}
\usepackage[table]{xcolor}
\usepackage[hidelinks,hypertexnames=false]{hyperref}
\usepackage{cleveref,caption,placeins}
\newcolumntype{Y}{>{\raggedright\arraybackslash}X}
\newcommand{\modelid}[1]{\texttt{\small #1}}
\title{Artificial Societies Benchmark:\\A Validation Framework for Synthetic Research}
\author{
    Edoardo Chidichimo\textsuperscript{\textdagger}\\%
    Artificial Societies\\%
    University of Oxford, UK\\%
    \And%
    Min Jun Jung\\%
    Artificial Societies\\%
    \And%
    Felix P. S. Wallis\textsuperscript{*}\\%
    Artificial Societies\\%
    \And%
    James K. He\textsuperscript{*}\\%
    Artificial Societies\\%
    \And%
    \texttt{[edo\textsuperscript{\textdagger},min,felix,james]@societies.ai}\\%
    {\small \textsuperscript{\textdagger}Corresponding author\quad \textsuperscript{*}Joint senior authors}
}

\begin{document}
\maketitle

\begin{abstract}
Synthetic surveys can reproduce average answers while misrepresenting how people differ, how their answers relate to each other, or how they respond to changes in conditions. We introduce the Artificial Societies Benchmark to help researchers assess whether synthetic populations support their intended analyses. The framework combines eleven tests across internal, construct, and external validity, drawing on twenty human sources and comparing nine language models. It connects each research use to its necessary evidence and tests how results change with information supplied about respondents. Importantly, strong performance in one domain does not establish fidelity in the others. Models often answer too consistently, compress response scales, and alter relationships between traits, while richer profiles improve prediction for some models and worsen it for others. The resulting scorecard helps researchers identify which aspects of a synthetic population can support their analysis and their work needs further human evidence.
\end{abstract}

\section{Introduction}
\label{sec:intro}
Suppose a synthetic survey reproduces the proportion of residents who support a proposed data centre expansion. A researcher might then use it to compare demographic groups, examine how support relates to concerns about energy use, or test how opinions vary with the proposal's expected benefits and costs. Predicting the overall level of support correctly does not tell us whether the simulation captures who supports the expansion, why they support it, what would change their views, or, perhaps more importantly, whether they would go out and sign a petition.

Human answers and behaviours should always be the basis for validating synthetic responses. Consider that people sometimes change their answer to the exact same question when researchers repeat it, and context can affect their responses. Achieving perfect consistency can therefore be a poor simulation of how people answer questions. Likewise, a model can produce plausible answers to individual survey questions while failing to preserve the relationships between what people report and what they do. These answers can support misleading conclusions about behaviour even when each answer sounds reasonable. Existing frameworks provide substantial evidence on selected requirements, but leave researchers to assemble these further checks themselves.

Our companion paper, \emph{Consequential Behaviour and Representational Fairness in the Validation of Synthetic Research} \cite{kutzner_consequential_2026}, sets out the broader standard for these claims. It gauges whether synthetic populations can predict what people do (often a more consequential endeavour than what people say), alongside checks of whether they reproduce human survey answers, variation between people, and relationships between answers. It also asks whether experimental interventions produce similar changes in synthetic and human responses while concurrently requiring evidence for the subgroups a decision affects. 

The present benchmark develops this diagnostic evidence through eleven tests we organise into three overlapping validity domains: (1) \textit{Internal validity} examines response processes, including how answers behave under repetition and changes to the survey;\footnote{We use internal validity to group response-fidelity diagnostics. Karetnikov and colleagues \cite{karetnikov_large_2026} use the term for causal identification in experiments on machines. Our EV-3 tests assess recovery of human experimental effects.} (2) \textit{Construct validity} examines psychological measurement, including relationships among questionnaire items, distinct traits, and independent criteria; and (3) \textit{External validity} examines population and experimental fidelity, including human response distributions, individual and subgroup differences, experimental effects, and the transfer of measurement structure to synthetic samples. In all, the benchmark brings tests of response processes, psychological measurement, and population and experimental fidelity into a common framework, helping researchers establish whether the evidence supports their intended analysis and render failures visible. Together, these papers explain what evidence a proposed use of synthetic research requires and how to test whether a synthetic population can adequately support it.

For demonstration, we implement the battery across twenty human sources and across nine large language models, comparing tests within the same respondents, instruments, and experimental assignments wherever the data allow. We also compare approaches to constructing personas and eliciting answers, vary the information we supply about respondents, and include simple statistical populations as controls. This design lets researchers examine whether apparent success on one test extends to the other requirements of their analysis, and whether richer personal information improves the resulting evidence.

Concretely, this paper contributes five elements:
\begin{itemize}
\item Eleven human-referenced tests connecting three validity domains to research uses.
\item Twenty human sources linking survey answers, experimental assignments, panel histories, and independent criteria.
\item Comparisons of simulation approaches, adapting Gordon and colleagues' protocol \cite{gordon_richer_2026}, to test whether demographic profiles, previous survey answers, and personal descriptions improve predictions of the same people's responses.
\item An evaluation of nine proprietary and open-weight large language models that retains estimates, uncertainty, and failures for each test.
\item Statistical population controls and human references that help explain what particular kinds of agreement establish.
\end{itemize}

\section{Background}

A recent review article \cite{karetnikov_large_2026} distinguishes four roles for language models as human proxies. \textit{Believable agents} portray convincing characters, \textit{task agents} perform work on people's behalf, \textit{experimental subjects} allow researchers to study the models themselves, and \textit{silicon samples} generate responses that researchers intend to support conclusions about humans. Our focus is the fourth role. 

In a synthetic survey, researchers give a model questions and information about a population or person, then analyse its answers as predictions of human responses. In practice, this might include prompting a model with a survey question whilst asking them to adopt a persona (based on some demographic information, for example). Richer personas can fold in interview transcripts or structured summaries to help enrich the model, in the hopes of more faithfully representing that archetypal persona. Others have experimented with the nature of the outputs, instructing models to emit probabilities across the available options rather than a forced single choice \cite{gordon_richer_2026}. Asking a model to simply estimate the proportion of a population likely to choose each possible answer to a survey question, either from the question alone or with information about the population's demographic composition, is another synthetic sampling method. Concretely, these choices determine both what the model knows about respondents and how their possible answers enter the resulting dataset.

Validation asks whether simulated responses reproduce particular features of human data. For example, researchers can compare population distributions, individual predictions, responses to experimental changes, or the relationships that make questionnaire scores meaningful. Park and colleagues \cite{park_llm_2026} build agents from participants' interviews and survey histories, then compare their predictions with answers those same participants gave to questions they withheld from the agents.\footnote{We retain the cited authors' terminology. Here, personas are the respondent descriptions we supply to models.} They benchmark survey prediction against participants' own agreement when answering again two weeks later and separately evaluate personality scores, economic-game choices, and experimental responses. SimBench \cite{hu_simbench_2026} tests whether models reproduce the proportions of people choosing each response across twenty datasets, measuring the distance between predicted and observed answer distributions, both for whole populations and demographic groups. SocioBench \cite{wang_sociobench_2025} gives models respondents' demographic information and asks them to predict their respective survey answers, scoring agreement with the recorded answers and examining how accuracy varies across countries, topics, and demographic groups. Jia and colleagues \cite{jia_when_2026} combine several checks using LISS (a Dutch panel that follows households over time), building personas from earlier survey records and testing their predictions against the same people's later answers. They assess individual accuracy, answer distributions, differences in accuracy across demographic groups, and whether synthetic responses recover the same clusters of respondents, finding that improved distributional agreement can coexist with weak individual prediction and distorted respondent groupings.

Other studies test whether synthetic data support the analyses researchers would conduct with human data. SSDataBench \cite{xie_evaluating_2026} compares synthetic and human datasets on five properties, covering individual variables' distributions, relationships between pairs of variables, prediction of outcomes from several variables, sequences of life events, and relationships between those sequences and personal characteristics. In other words, its statistical tests assess whether the generated data reproduce these population patterns. Ashokkumar and colleagues \cite{ashokkumar_large_2026} give models the original experimental materials and demographic profiles, generate responses for each experimental condition, and calculate the resulting treatment effects. They compare these with human effects across experiments, examining both whether larger human effects receive larger predictions and whether the predicted magnitudes are accurate. Their finding that predictions correlate strongly with human effects while systematically overestimating their size illustrates why, for example, these ought to remain separate validation questions. 

A further, psychometric validation question concerns whether synthetic answers retain the relationships that make a questionnaire useful for measuring a psychological characteristic. This requires checking how questionnaire items relate to one another, whether they form the same underlying dimensions, and whether scores have comparable meanings across human and synthetic samples. Lukauskas and \v{S}arkauskait\.{e} \cite{lukauskas_plausible_2026} examine these properties in questionnaires measuring attitudes towards organisational change, work engagement, and job performance. They assess scale reliability, underlying dimensions, and relationships between measures, finding that synthetic data can distort these relationships and produce statistically significant indirect effects where the human estimates do not reach significance. Shen and Han \cite{shen_reliability_2026} apply factor analysis and measurement-invariance tests to depression and anxiety questionnaires. They find high internal consistency alongside poor fit to the human factor structure and failed scalar invariance. Karetnikov and colleagues \cite{karetnikov_large_2026} place validation within the broader question of what researchers intend a simulation to predict. They distinguish directional agreement from accurate magnitudes, require evaluation against held-out human data, and examine whether agreement extends to new populations, times, and settings.

Together, these studies establish complementary ways to validate synthetic responses, from population distributions and individual predictions to experimental effects and psychological measurement. In line with Jia and colleagues' multidimensional evaluation \cite{jia_when_2026}, our benchmark extends this approach by organising response-process, measurement, and population comparisons around the evidence that a proposed research use requires, so that researchers can identify both the checks they can perform and the claims their data cannot yet support.

\section{The Artificial Societies Benchmark}
\label{sec:benchmark}

The Artificial Societies Benchmark organises eleven tests into three overlapping validity domains, covering response processes under internal validity, psychological measurement under construct validity, and population and experimental fidelity under external validity. Consider an electricity supplier deciding whether a cheaper overnight tariff will persuade households to shift electric-vehicle charging, as in our companion framework's example \cite{kutzner_consequential_2026}. A synthetic survey might reproduce average willingness to switch while confusing financial incentives with access to a home charger, overlooking renters' constraints, or exaggerating the response to a price change. These errors would affect whom the supplier expects to benefit and how much demand it expects to shift. Researchers therefore need tests of response processes, relationships, groups, and experimental effects; a claim about actual charging additionally needs comparison with observed charging behaviour.

Table~\ref{tab:measures} summarises the tests, and the descriptions below explain their motivation, measures, and interpretation. Appendix~\ref{app:metrics} supplies the formal definitions.

\begin{table}[!htbp]
\centering
\caption{\textbf{What each test asks about a synthetic population.} Use the intended analysis to select relevant tests across the three domains. Appendix~\ref{app:metrics} gives formal definitions.}
\label{tab:measures}
\begingroup
\setlength{\tabcolsep}{4pt}
\renewcommand{\arraystretch}{1.15}
\begin{tabularx}{\textwidth}{@{}lYYY@{}}
\toprule
Test & Measure & Comparison & What it captures\\
\midrule
\multicolumn{4}{@{}l}{\textit{Internal validity}}\\
IV-1 & Repeated questions & Repeat the same item & Excess consistency or variability\\
IV-2 & Linked questions & Check logical constraints & Departures from human violation patterns\\
IV-3 & Paraphrased questions & Ask equivalent wording & Changes in answers across wording\\
IV-4 & Perturbed surveys & Change survey administration & Human sensitivity to the instrument\\
\midrule
\multicolumn{4}{@{}l}{\textit{Construct validity}}\\
CV-1 & Item covariance & Relate items within a scale & Coherence of the measured construct\\
CV-2 & Independent criteria & Relate answers to held-out criteria & Relationships beyond the target items\\
CV-3 & Distinct constructs & Compare within- and between-trait links & Separation among measured traits\\
\midrule
\multicolumn{4}{@{}l}{\textit{External validity}}\\
EV-1 & Response distributions & Compare option frequencies or values & Population-level response fidelity\\
EV-2 & Groups and individuals & Compare groups or linked people & Conditional and person-level fidelity\\
EV-3 & Experimental effects & Replay assigned treatment contrasts & Direction and magnitude of response\\
EV-4 & Latent structure & Fit the same measurement structure & Loadings, correlations, and invariance\\
\bottomrule
\end{tabularx}
\endgroup
\end{table}

\subsection{Internal validity}

\paragraph{IV-1 Repeated questions.} A model may give the same answer every time researchers repeat a question, even when the represented people change their answers on a later occasion. We compare human and synthetic repeat agreement using exact agreement, Cohen's kappa, which adjusts for agreement we expect from response frequencies, and changes in numerical answers \cite{cohen_coefficient_1960,park_llm_2026}. Agreement above the human reference indicates excess consistency, agreement below it indicates excess variability, and researchers should likewise interpret numerical changes against human retest variation.

\paragraph{IV-2 Linked questions.} A respondent may assign a higher probability to two events occurring together than to either event alone, while a model that never makes such errors may also misrepresent human judgement \cite{sundh_unified_2023}. We measure how often and by how much answers violate specified logical constraints, and compare the combinations of violations with those we observe in human answers. Smaller human--synthetic differences indicate closer reproduction of these response patterns, while either more or fewer violations can signal a mismatch.

\paragraph{IV-3 Paraphrased questions.} A model may change its answer when researchers rephrase a question without changing its meaning, a sensitivity researchers examine using ValueConsistency (a dataset of value-laden questions and alternative wordings) \cite{moore_are_2024}. We measure agreement across equivalent wordings and the distance between human and synthetic answer-transition distributions, which record how often each original answer changes to each alternative. Closer agreement with the human reference and smaller transition distances indicate more faithful wording sensitivity, with excessive stability also counting as a departure.

\paragraph{IV-4 Perturbed surveys.} Models can fail to reproduce familiar human response biases and react to wording changes that people generally tolerate \cite{tjuatja_llms_2024}. We compare the direction and size of human and synthetic response shifts under changes to wording, order, labels, and other survey features, using differences in mean shifts or changes in response-option frequencies. Errors near zero indicate that human and synthetic responses show similar sensitivity to survey design. Larger errors indicate differences that could affect the study’s conclusions.

\subsection{Construct validity}

\paragraph{CV-1 Item covariance.} Plausible answers to individual questionnaire items can conceal associations that are too weak or too strong for the items to function together as they do in humans \cite{lukauskas_plausible_2026}. We compare item associations and scale reliability, including Cronbach's alpha and McDonald's omega, with their human values to detect altered relationships within a proposed construct \cite{cronbach_coefficient_1951,mcdonald_test_1999}. Smaller discrepancies indicate closer within-scale measurement patterns, while unusually high reliability can reflect excessive similarity between answers and does not by itself establish validity.

\paragraph{CV-2 Independent criteria.} A synthetic questionnaire score may appear coherent while losing the relationship that human scores have with an independently measured characteristic or outcome, undermining the evidence connecting the measure to its intended construct \cite{cronbach_construct_1955}. We compare how strongly human and synthetic questionnaire scores relate to the same independent measure for the same participants. Errors near zero indicate that synthetic responses preserve the relationship observed in the human data. The independent measure determines what conclusions that relationship supports.

\paragraph{CV-3 Distinct constructs.} A model may make distinct characteristics, such as extraversion and conscientiousness, so closely related that their scores become difficult to distinguish, weakening discriminant validity \cite{campbell_convergent_1959}. We compare correlations between trait scores and the separation between within-trait and between-trait item associations against the corresponding human relationships. Smaller discrepancies indicate that the synthetic data distinguish constructs similarly to the human data, while departures in either direction can distort analyses involving several traits.

\subsection{External validity}

\paragraph{EV-1 Response distributions.} A model can reproduce an average while concentrating answers near the centre and missing people at either extreme, a problem researchers document in synthetic survey data \cite{bisbee_synthetic_2024}. We compare human and synthetic answer distributions using total variation distance for categorical responses, as in SimBench \cite{hu_simbench_2026}, and range-normalised Wasserstein distance for numerical responses, capturing discrepancies in option frequencies and the displacement of values across the scale. Smaller distances indicate closer population distributions, but leave open whether models assign answers to the right people or retain the right relationships.

\paragraph{EV-2 Groups and individuals.} An accurate population distribution can conceal poor predictions for particular demographic groups or for the people whose profiles generated the answers \cite{jia_when_2026}. We compare distributions within demographic groups and, where we link records, calculate each person's proportion of incorrect categorical answers or average numerical error relative to the response range. Lower errors indicate closer agreement for the evaluated groups or people, while groups with insufficient human evidence remain unresolved.

\paragraph{EV-3 Experimental effects.} A model may predict that an intervention changes responses in the right direction while substantially exaggerating its effect, as researchers observe in experimental prediction research \cite{ashokkumar_large_2026}. We calculate the treatment--control difference in human and synthetic responses, then report the signed and absolute difference between those effects. Smaller absolute errors indicate closer effect recovery, while the signed error identifies the direction of the discrepancy and the paired effects reveal attenuation, exaggeration, or reversal.

\paragraph{EV-4 Latent structure.} Synthetic answers can appear highly consistent while departing from the human measurement structure or using response categories differently, preventing equivalent score interpretation \cite{shen_reliability_2026}. We fit the same measurement structure to human and synthetic data, compare item loadings and factor correlations, and use ordinal measurement-invariance tests to examine whether response thresholds and item--factor relationships are comparable. Smaller structural discrepancies and supported invariance strengthen comparisons on the fitted dimensions, while unused response categories can prevent estimation and remain explicit evidence of scale collapse.

\begin{figure}[!htbp]
\centering\includegraphics[width=\textwidth]{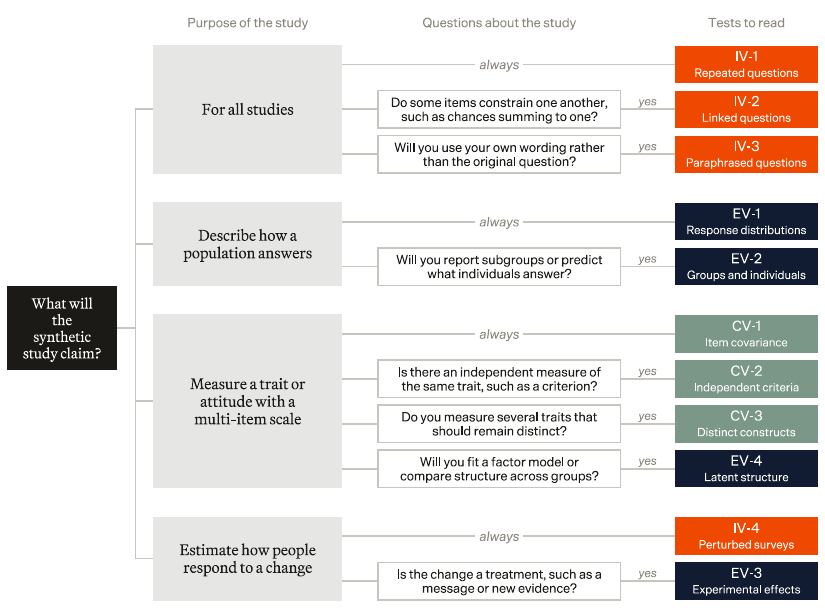}
\caption{\textbf{Choosing tests for a proposed study.} Following relevant branches shows which tests a study requires. The top branch applies across study purposes. Colours identify the three validity domains.}
\label{fig:which_tests}
\end{figure}

For ease, we provide Figure~\ref{fig:which_tests} to guide researchers in their selection of tests for their proposed claim(s), where analysis should entail reading their estimates alongside human references, uncertainty, and coverage. We purposefully encourage researchers to produce scorecards based on these tests (at least on those relevant to the research question) and we avoid imposing a universal pass threshold. This way, we do not fold failures and unavailable estimates into a single score and researchers can themselves evaluate which model candidates perform best for their use-case whilst cognisant of any emergent discrepancies that these metrics flag. Formal equivalence claims additionally require prespecified, use-specific tolerances and uncertainty in the human--synthetic comparison.

\section{Methods}
\label{sec:methods}
\begin{figure}[!htbp]
\centering\includegraphics[width=\textwidth]{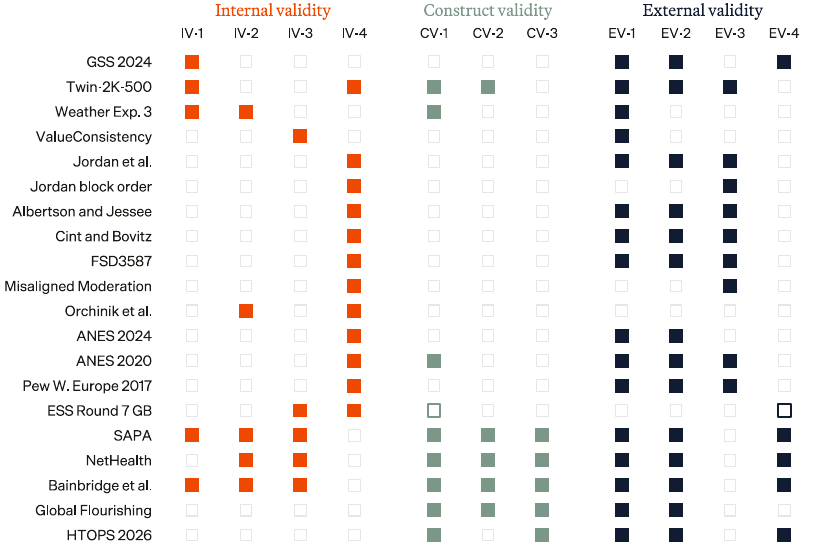}
\caption{\textbf{Which datasets yield estimates for each test?} Filled squares indicate at least one model yields an estimate, coloured outlines indicate a test we attempted without an estimate, and pale outlines indicate a combination we did not run. Counts appear in Table~\ref{tab:counts}.}
\label{fig:coverage}
\end{figure}

\paragraph{Human datasets.} The Artificial Societies Benchmark draws on twenty human data sources, comprising national surveys, panels that follow the same people over time, personality inventories, and experiments. These datasets contain recorded answers from real participants and, depending on the study, demographic profiles, earlier survey responses, written self-descriptions, experimental assignments, or independently measured characteristics and outcomes. Repeated questions and alternative survey versions reveal response consistency and sensitivity to wording or order, while multi-item scales and independent measures support checks of psychological measurement. Recorded answers provide reference distributions for populations and subgroups, linked records allow predictions for the same people, and experimental assignments support treatment effect comparisons. Figure~\ref{fig:coverage} shows which datasets support each test: a filled square means that the dataset yields an estimate for at least one model and a coloured outline means that we attempted the test but could not estimate it. To be clear, an estimate can show either close agreement or a large error; the filled square alone does not indicate success. Appendix~\ref{app:design} describes each dataset and its simulation inputs.

\paragraph{Models.} We select nine state-of-the-art models spanning proprietary and open-weight systems. Table~\ref{tab:models_summary} summarises their weight availability and generation settings (see Table~\ref{tab:models} for exact model identifiers, access routes, and additional decoding settings).

\begin{table}[!htbp]
\centering\small
\caption{\textbf{Models and generation settings.} Open-weight denotes public availability of model weights. Gemini's backend ignores temperature; Scout has no thinking mode.}
\label{tab:models_summary}
\begin{tabularx}{\textwidth}{@{}Y l l l@{}}
\toprule
Model & Weight availability & Temperature & Thinking \\
\midrule
GPT-5.6 Sol & Proprietary & 1 & Off \\
Claude Opus 5 & Proprietary & 1 & Off \\
Gemini 3.8 Flash & Proprietary & Ignored & Low \\
Grok 4.3 & Proprietary & 1 & Off \\
Mistral Small 2603 & Open-weight & 1 & Off \\
DeepSeek Flash & Open-weight & 1 & Off \\
Qwen3.8-27B & Open-weight & 1 & Off \\
Gemma 4 31B & Open-weight & 1 & Off \\
Llama 4 Scout & Open-weight & 1 & None \\
\bottomrule
\end{tabularx}
\end{table}

\paragraph{Simulation and comparison.} Most tasks include up to 512 people, with source-specific exceptions. Linked records stay together when assigning development and evaluation samples. Models receive the source question and available profile information, while we withhold target answers; sequence tasks also retain earlier generated answers. The resulting synthetic respondents do not interact. We also use two statistical controls that generate answers without a language model: an \emph{independent-marginal} control samples each item separately from its human response distribution, while a \emph{Gaussian rank-copula} control approximates relationships between items. Following the use of such controls in psychometric validation \cite{lukauskas_plausible_2026}, we use them to distinguish matching individual answer frequencies from preserving joint response patterns; Appendix~\ref{app:baselines} gives our implementation.

\paragraph{Simulation protocol.}
\label{sec:prolific_protocol}
We adapt Gordon and colleagues' Prolific protocol \cite{gordon_richer_2026} to compare how the information and answer format researchers give a model affect survey prediction. Their protocol contrasts providing only questions, questions with population demographics, individual demographic profiles, and profiles with personal accounts. It also compares original accounts with structured summaries and alternative ways to request answers. We implement the profile types supported by each source, using written self-descriptions in place of interviews and adding earlier survey histories where available. We do not reproduce the protocol's full set of reasoning, calibration, and per-person probability-output experiments.

\paragraph{Comparing simulation approaches.}
\label{sec:elicitation}
In the General Social Survey (GSS, a US survey of social attitudes), we implement the protocol's first three information conditions using the GSS questions and available demographics. Models either predict population answer proportions from questions alone, predict those proportions with demographic composition, or answer individually from demographic profiles. The first two approaches return percentages across options; the third returns individual answers that we aggregate. Twin-2K-500 (Twin), a four-wave panel with survey histories, personal descriptions, and behavioural questions, supports the demographic, original-text, and distilled-text conditions, plus our extension that adds earlier survey answers to each. We compare these six conditions on the same people and target questions, measuring prediction error from zero to one and averaging within person. Lower error means closer agreement with the person's recorded answer.

\paragraph{Checking structured profiles.}
\label{sec:profile_audit}
A shared preprocessing model, \modelid{gpt-5.4-2026-03-05}, organises Twin's personal descriptions into seven fields: values, moral foundations, decision style, institutional trust, political engagement, key concerns, and consumer orientation. We call these summaries distilled profiles. A second model, \modelid{claude-opus-5}, checks whether each statement in the generated profile follows from the person's original answers, including whether it correctly distinguishes current characteristics from aspirations. Our scripts then check quotations against the source text and screen for target-answer text. These checks assess only whether the profile represents the source faithfully, not its subsequent prediction accuracy. We retain the 430 of 512 profiles that pass and compare all six conditions on those same people, using identical profiles across the nine models. Appendix~\ref{app:profile_audit} details the checks and exclusions.

\paragraph{Scoring and uncertainty.}
\label{sec:scoring}
We first check whether enough usable model answers and corresponding human answers exist to support a comparison. For most estimates, our scorecard requires 90 percent valid model answers overall and on each item, 90 percent coverage of the human-observed targets, with at least 30 people. Exploratory structural comparisons apply this coverage threshold among humans with complete battery data and require at least 100 complete matched participants. These operational thresholds limit reporting from sparse or incomplete outputs, though they are not universal requirements or guarantees of validity. We estimate uncertainty by resampling people 500 times, keeping their answers together and preserving experimental assignments.

Twin's paired information comparisons use the same 430 eligible people but compare only questions on which both conditions give valid answers, with at least 30 contributing people and item coverage alongside each estimate. To assess whether unanswered questions could alter the conclusion, we also calculate the most and least favourable full-question-set differences that their unknown errors permit. Appendix~\ref{app:twin_sensitivity} defines this sensitivity analysis. Every model follows the same parsing and scoring rules, documented in Appendix~\ref{app:sources}.

\section{Results}
\label{sec:results}

\begin{figure}[!htbp]
\centering\includegraphics[width=\textwidth]{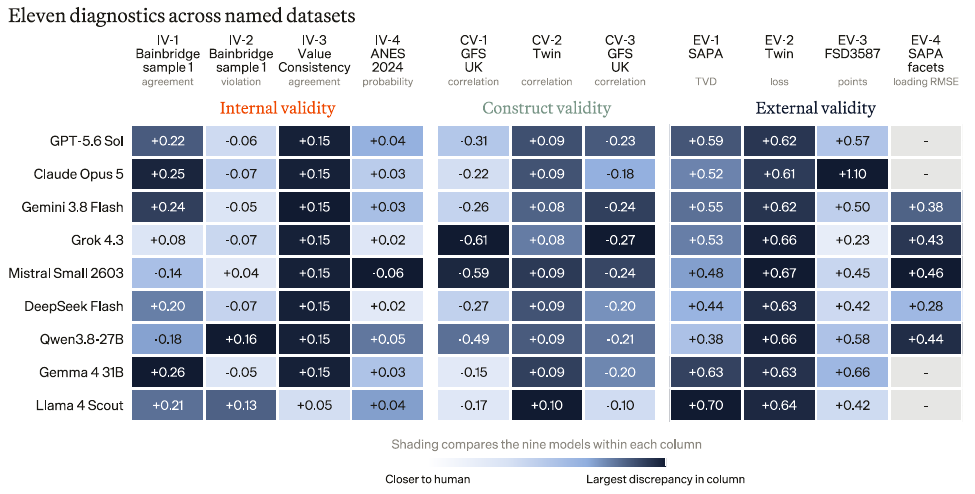}
\caption{\textbf{Model departures from human responses across eleven validity tests.} Each row represents a model, and each column shows one diagnostic from the dataset named in its heading, using the first generation. Cell values report differences or prediction errors in the units of that diagnostic. For signed differences, positive values mean that the model statistic exceeds the human statistic, and negative values mean that it falls below it. For example, $+0.22$ in IV-1 means that the model repeats its answers 22 percentage points more often than humans. Absolute errors and prediction losses are nonnegative. Across all columns, values closer to zero indicate closer agreement or lower error. White represents zero; darker shading indicates larger absolute values relative to the largest among the nine models within that column. Each column has its own units and colour scale, so comparisons apply within columns. CV-1 and CV-3 use the UK Global Flourishing Study (GFS). EV-4 measures error in how strongly SAPA facet items relate to their traits. A dash marks an unavailable estimate. Appendix~\ref{app:scorecard_guide} defines each diagnostic and explains unavailable estimates.}
\label{fig:scorecard}
\end{figure}

Our eleven tests reveal distinct failures across models and sources. Table~\ref{tab:counts} reports study-family coverage for each test. Figure~\ref{fig:scorecard} illustrates model performance using one diagnostic and dataset per test. We report results separately for each dataset to preserve differences in what the tests measure.

\subsection{Reliable-looking answers can hide measurement problems}
\begin{figure}[!htbp]
\centering\includegraphics[width=\textwidth]{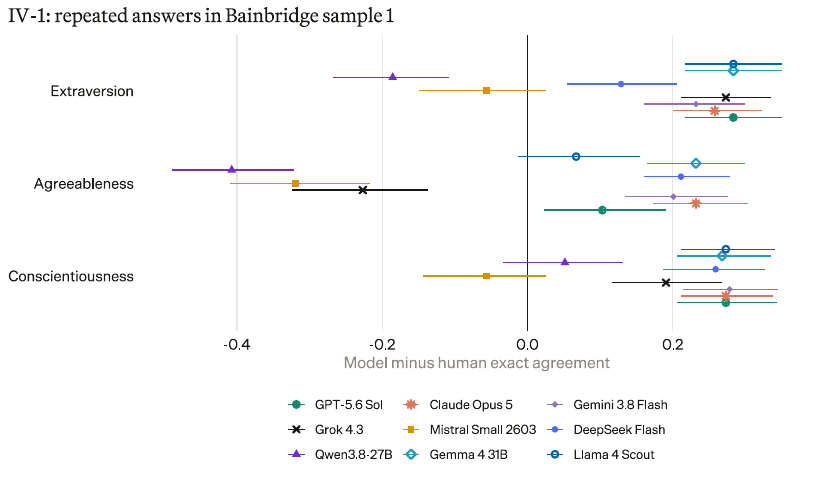}
\caption{\textbf{IV-1: repeated-answer agreement in Bainbridge sample 1.} Each point compares a model with humans on one of three identically worded item pairs. Zero means equal repeat agreement, positive values mean that the model repeats itself more often, and negative values mean that it repeats itself less often. Lines show paired 95 percent participant-bootstrap intervals. Matching human agreement, rather than maximising consistency, is the comparison of interest.}
\label{fig:retest}
\end{figure}

Models differ from humans in repeated-answer agreement and responses to opposing personality items (IV-1, IV-2). In Bainbridge sample 1, humans give the same answer to duplicate personality items about 73 percent of the time, compared with 54 percent for Qwen and 99 percent for Gemma; seven models exceed the human average (Figure~\ref{fig:retest}). Scout repeats answers about 93 percent of the time yet gives contradictory responses to opposing personality items about 21 percent of the time. The contradiction check counts answers on the same side of the scale midpoint for opposite statements, such as endorsing both a characteristic and its opposite; human contradictions average about 8 percent. This scoring rule flags opposing endorsements, which need not imply a logical impossibility because behaviour can vary by context. Scout therefore shows both higher repeated-answer agreement and more opposing endorsements than humans.

Models also smooth away human variation across equivalent wording and logically related questions (IV-3, IV-2). In ValueConsistency, humans agree across paraphrases about 85 percent of the time, Scout about 90 percent, and the other eight models 100 percent. In Orchinik and colleagues' consensus task, respondents state their belief under hypothetical scientific agreement of 50, 75, 90, 97, and 99 percent. About 48 percent of human response sequences contain at least one decrease in belief as agreement rises, whereas no model sequence does. Both tasks show less response variability in models than in humans.

Models differ in how closely they reproduce human option-order effects (IV-4). In ANES 2024, reversing the response options on an attention-to-politics question reduces the human share choosing the first coded category by about 4.4 percentage points. The corresponding first-generation change is about $-0.3$ points for OpenAI, zero for Scout, and $+0.5$ points for Qwen, while Mistral changes by about $-10.6$ points. These estimates track the same substantive category across the two orders. The estimated order effects vary in both magnitude and direction across models.

Missing response categories and constant answers prevent some comparisons of relationships and latent structure (CV-1, EV-4). Every model omits ESS response categories that humans use, despite an adequate reference fitted on 397 development participants. For example, humans use all eleven positions on an immigration scale, while OpenAI uses only 6 to 10 on that item. Our ordinal comparison retains the original response categories and requires observations in each category to estimate the boundaries between adjacent options. Gemma also gives every person the same probability on 28 of 60 weather-judgement items, leaving no variation with which to estimate their rank associations (CV-1). These responses provide insufficient variation for the specified comparisons.

SAPA's well-being, sensation-seeking, and honesty facets provide numerical evidence about item--trait structure (EV-4). Five models yield loading-error estimates from 0.28 to 0.46; smaller values mean that items relate to their traits more similarly to the human reference. The other four models lack a usable estimate because responses or fits do not support the comparison. These exploratory results quantify structural differences where estimation succeeds, without establishing measurement invariance.

\begin{figure}[!htbp]
\centering\includegraphics[width=\textwidth]{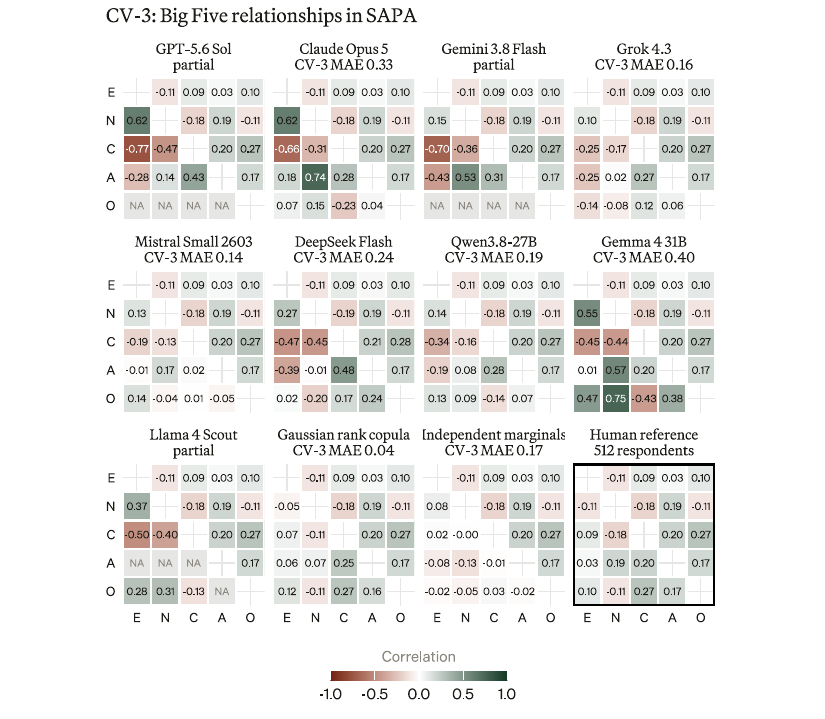}
\caption{\textbf{CV-3: Big Five trait relationships in SAPA.} Each model or control panel places synthetic correlations below the diagonal and correlations for its matched humans above it. The black-bordered bottom-right panel shows the full 512-person human reference. E, N, C, A, and O denote extraversion, neuroticism, conscientiousness, agreeableness, and openness. Compare corresponding cells: agreement with the human pattern is desirable rather than high correlation across the matrix. Panel subtitles give mean absolute correlation error against the matched humans; incomplete summaries say ``partial''. NA indicates a constant synthetic trait with no defined correlation. The two statistical controls isolate the contribution of item dependence.}
\label{fig:structure}
\end{figure}

The two statistical controls differ in how closely they reproduce between-trait relationships (CV-3, Figure~\ref{fig:structure}). In SAPA, an online personality dataset, the Gaussian rank-copula and independent-marginal controls both reproduce individual item distributions closely (EV-1), but their mean absolute errors in correlations between traits are 0.040 and 0.170 respectively. The latter means an average discrepancy of 0.17 correlation units, more than four times the former. The independent control draws each item separately and yields larger errors in trait relationships despite close agreement on item distributions.

\subsection{The strongest model depends on the research question}
\begin{figure}[!htbp]
\centering\includegraphics[width=\textwidth]{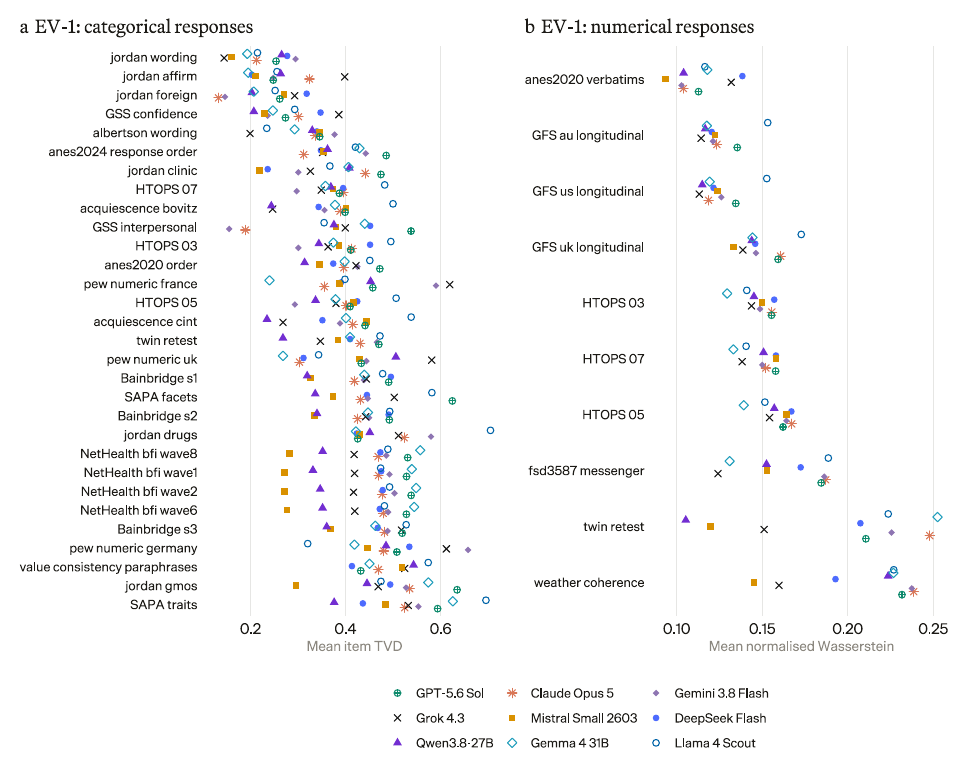}
\caption{\textbf{EV-1: response-distribution error by dataset and model.} Points show average item-level error for the first generation, with smaller values indicating closer human agreement. The left panel compares categorical answer frequencies using total variation distance; the right compares numerical responses using Wasserstein distance divided by the response range. Both have zero at exact agreement, but they measure different discrepancies. Only tasks with complete estimable item support enter these means.}
\label{fig:marginals}
\end{figure}

Success at reproducing response distributions can coexist with unusual responses to opposing personality statements or weak separation between constructs (EV-1, IV-2, CV-3, Figure~\ref{fig:marginals}). Qwen has the smallest mean distributional error on the first SAPA Big Five generation, at 0.38 total variation distance, but the highest contradiction rate in Bainbridge sample 1, at 0.24 against the human rate of 0.08. A distance of 0.38 means that, on average across items, 38 percent of probability mass would need to move between answer options to match the human proportions. In March HTOPS, the US Census Household Trends and Outlook Pulse Survey, DeepSeek, Grok, Mistral, and Qwen yield supported gaps between within-construct and between-construct item correlations from $-0.09$ to $0.07$, compared with about 0.20 for humans. Constant responses prevent complete estimates for OpenAI, Claude, and Gemma. Gemini and Scout have too few finite bootstrap estimates. A small gap means that items about different constructs relate almost as strongly as items intended to measure the same construct.

Subgroup estimates depend on the number of respondents available for each item and group (EV-2). The Cint and Bovitz acquiescence experiments test whether people tend to agree with statements regardless of whether their wording expresses a proposition or its opposite. Both support those experimental comparisons, but their demographic groups contain too few respondents per item for most subgroup checks. Cint has only one eligible item--group combination out of 400, and Bovitz has none. Subgroup accuracy remains largely unassessed in these two sources.

\begin{figure}[!htbp]
\centering\includegraphics[width=\textwidth]{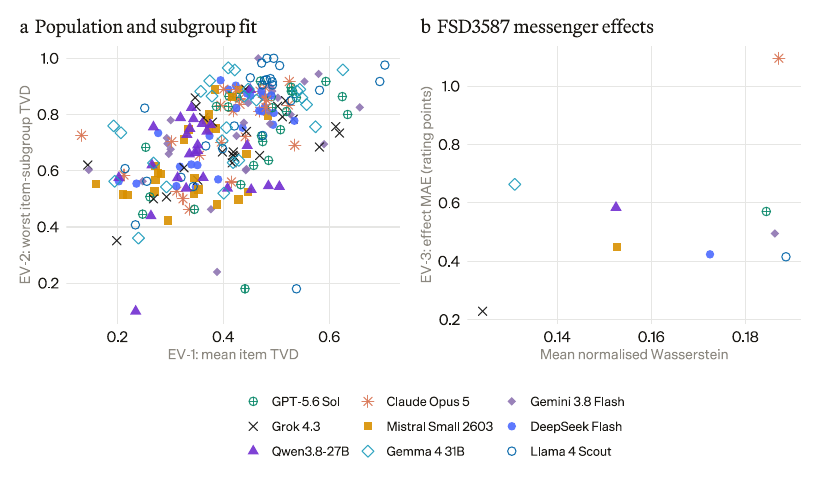}
\caption{\textbf{What overall response agreement leaves unresolved.} Each point represents one model and task in the first generation. Left (EV-1, EV-2), smaller horizontal values mean closer overall answer frequencies, and smaller vertical values mean closer agreement for the worst eligible item--subgroup combination; the lower-left corner indicates smaller errors on both. Right (EV-1, EV-3), FSD3587 points compare response-distribution error with mean error in the effect of changing the messenger, in 0--10 rating points. Again, smaller values on both axes indicate closer human agreement. The panels use different distance measures and do not establish statistical independence between tests.}
\label{fig:dissociation}
\end{figure}

\begin{figure}[!htbp]
\centering\includegraphics[width=\textwidth]{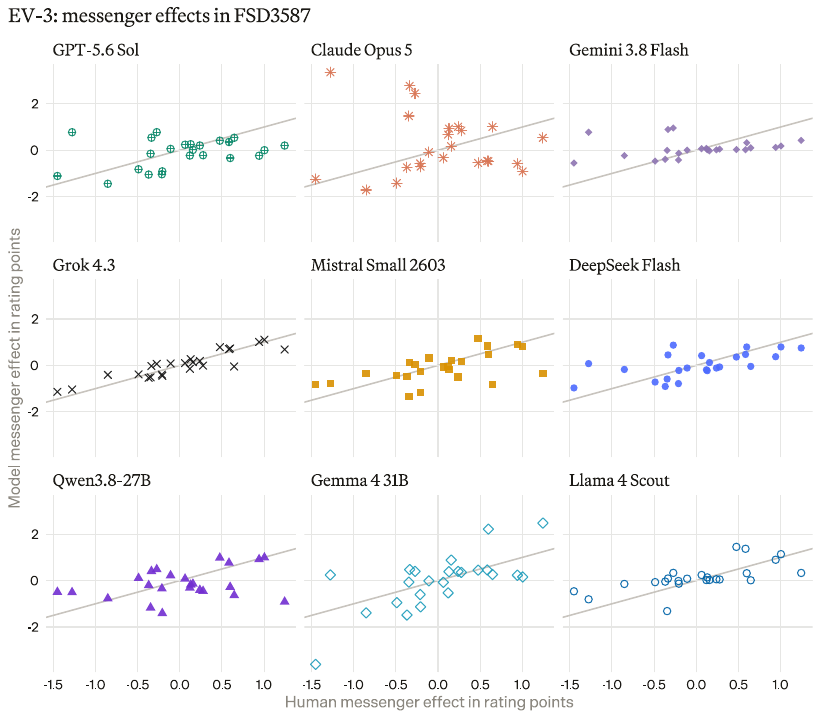}
\caption{\textbf{EV-3: messenger-effect recovery in FSD3587.} Each panel shows 24 contrasts on the 0--10 argument-rating scale. Points on the diagonal match the human effect; departures show its attenuation, amplification, or reversal. Values with opposite signs indicate reversal. The scorecard retains model-minus-human error intervals for each contrast.}
\label{fig:effects}
\end{figure}

Error in experimental effects varies across models (EV-3, Figure~\ref{fig:dissociation}). Across 24 FSD3587 contrasts that vary the stated messenger of a policy argument, Grok has the smallest mean absolute effect error, at 0.23 rating points, and Claude the largest, at 1.10, against an average absolute human effect of 0.52 (Figure~\ref{fig:effects}). For several models, the average error is as large as the human effect itself. Individual contrasts show attenuation, exaggeration and reversal of human effects.

Average correlation errors conceal larger discrepancies for specific item--criterion pairs (CV-2). Across 108 Twin model--condition--repetition summaries, mean absolute error in correlations with independent criteria is about 0.08 to 0.10. Each average covers 168 to 256 estimable pairs, so differences can reflect which pairs qualify as well as prediction quality. These criteria include numeracy and personality measures that do not enter the prompts. For an Allais-choice item, the human association with numeracy is about 0.28, whereas OpenAI with demographic profiles yields about 0.11, a loss of 0.17 correlation units. For this item, OpenAI with demographic profiles retains less than half the human association with numeracy.

\subsection{More personal information helps selectively}
\begin{figure}[!htbp]
\centering\includegraphics[width=\textwidth]{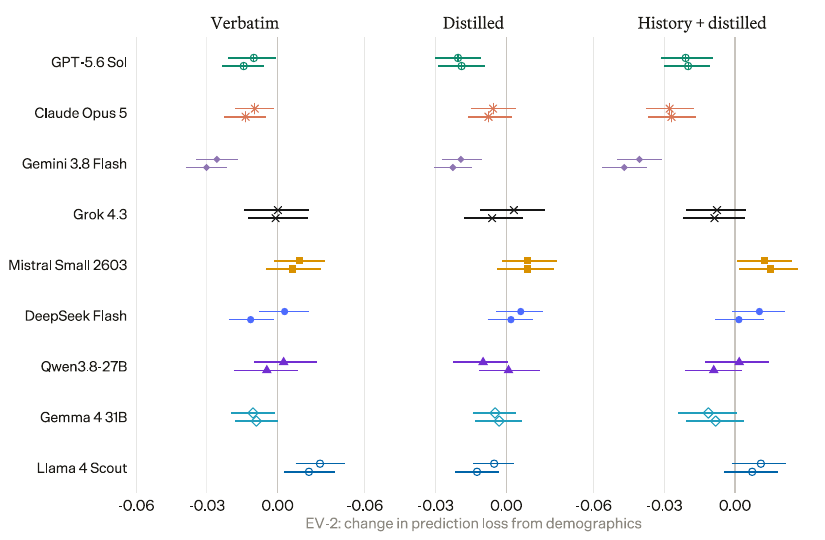}
\caption{\textbf{EV-2: does personal information improve prediction over demographics alone?} Each point gives the change in average prediction error for one model and repetition among the same 430 people. Points left of zero indicate improvement, and points right of zero indicate worsening; the two points per row represent separate generations. Lines show paired 95 percent participant-bootstrap intervals, and an interval crossing zero leaves the direction uncertain. The three panels add original self-descriptions, distilled descriptions, or survey history plus distilled descriptions. These primary comparisons use jointly valid answers; Table~\ref{tab:twin_sensitivity} also accounts for unanswered questions.}
\label{fig:ladder}
\end{figure}

The Twin comparisons test whether richer profiles improve predictions for the same individuals (EV-2). All 270 comparisons cover the common cohort of 430 people, with thirty comparisons per model across two repetitions, and Figure~\ref{fig:ladder} shows changes relative to demographic profiles. We compare the same person and question wherever both conditions produce a valid answer, then average within person. A decrease of 0.02 means two hundredths less error on the zero-to-one loss scale, not a 2 percent relative increase in accuracy. These comparisons describe changes in prediction for people whose distilled profiles pass the audit; the sensitivity analysis also accounts for differences in unanswered questions.

Distilled profiles improve individual prediction for some models and worsen it for others (EV-2). OpenAI and Gemini reduce error by about 0.02 in both repetitions, with intervals excluding zero. History plus distilled text reduces error relative to demographics alone by 0.047 and 0.041 for Gemini and 0.027 and 0.028 for Claude, while Mistral's error increases by 0.015 and 0.012, with both primary intervals excluding zero. Grok and DeepSeek show smaller, uncertain changes (Appendix~\ref{app:ladder_results}). The direction and size of the change depend on the model and profile condition.

The missing-answer sensitivity analysis supports the main improvements but weakens the evidence for Mistral's worsening (EV-2). We hold observed errors fixed and allow each unanswered question any error from zero to one, then average over the full human-observed question set for each person. The bootstrap envelopes remain below zero for OpenAI and Gemini with distilled profiles and for Claude and Gemini with history plus distilled profiles in both repetitions. Mistral's history-plus-distilled sample bounds remain positive, but its envelopes extend from approximately $-0.005$ to $0.032$ and $-0.005$ to $0.030$, including zero. The direction of the estimated worsening survives the missing-answer bounds in this sample, while the combination of missingness and sampling uncertainty prevents a firm directional conclusion.

Scout's individual prediction error differs between original and distilled profiles (EV-2). Scout's original self-descriptions increase error relative to demographics, whereas distilled descriptions reduce its point estimates. Directly comparing distilled with original text gives reductions of 0.026 and 0.023 across repetitions, with both sensitivity envelopes below zero. Distilled profiles therefore yield lower error than original descriptions for Scout in both repetitions.

Item-level failures also limit the available measurement estimates (CV-1 to CV-3, EV-4). For SAPA's personality item ``Tell the truth'', Gemma fails to give a rating in 478 of 512 requests, instead asking for a phrase to rate; DeepSeek declines 39 times. Gemma appears to treat the item as an instruction. The appendix reports these failures alongside sequence completion. For Gemma, failures on this item prevent the facet comparison.

\section{Discussion}
\label{sec:discussion}
The benchmark shows why validation must follow the analysis a researcher intends to perform. Work such as SimBench and SocioBench establishes ways to assess population distributions and individual predictions, while Jia and colleagues show that improvement on one dimension can coexist with deterioration on another \cite{hu_simbench_2026,wang_sociobench_2025,jia_when_2026}. Our results extend that concern to response processes and psychological measurement within a common testing framework. Qwen's relatively close SAPA distributions accompany frequent contradictions in Bainbridge, and all nine models omit ESS response categories that humans use. The scorecard connects each proposed use of a synthetic sample to the evidence needed to support it.

Human response variability will ultimately give these diagnostics their interpretation. Park and colleagues compare survey prediction with human retest agreement, while Tjuatja and colleagues examine how models respond to changes in survey administration \cite{park_llm_2026,tjuatja_llms_2024}. Our duplicate-item and paraphrase findings show why more consistency can mean less human fidelity, and the ANES order comparison shows that models can also miss the size or direction of a human method effect. Simulating questionnaire responses requires preserving the variation humans show under repetition and changes to wording or order. These tests assess observable response patterns and leave the underlying cognitive mechanisms unresolved.

Psychometric fidelity determines whether plausible item answers support meaningful comparisons of traits. Lukauskas and \v{S}arkauskait\.{e} and Shen and Han show that internally consistent synthetic responses can distort measurement structure \cite{lukauskas_plausible_2026,shen_reliability_2026}. Our trait matrices and HTOPS construct gaps expose related problems, while the independent-marginal control demonstrates how close item frequencies can coexist with weak preservation of trait relationships. A researcher studying links between well-being and institutional trust therefore needs evidence about those constructs and their relationships, beyond agreement on their average scores. ESS category collapse also restricts the measurement evidence available. It demonstrates limited scale use but leaves measurement invariance unassessed.

Simulation construction is part of what researchers must validate. Gordon and colleagues' protocol separates the information a model receives from the way researchers elicit its answers \cite{gordon_richer_2026}. Our adaptations show that the benefit of additional information depends on both model and representation. OpenAI and Gemini improve with distilled text, while Scout responds differently to original and distilled descriptions. The sensitivity analysis supports those comparisons but qualifies Mistral's worsening once missing answers and sampling uncertainty both enter the assessment. All models receive identical distilled profiles, and the use of OpenAI for distillation and Claude Opus 5 for auditing is a limitation because it may favour models from those providers. Researchers should therefore compare the exact profile and elicitation procedure they intend to use, report unsuccessful answers alongside accuracy, and check whether exclusions change the population or questions to which their claim applies.

Our companion framework distinguishes predicting what people say from predicting what they do \cite{kutzner_consequential_2026}. Its four diagnostic levels ask whether synthetic populations preserve typical responses, variation between people, sensitivity to context, and relationships among measures. The eleven tests here provide practical comparisons across those properties. Its behavioural criterion instead asks whether predictions agree with consequential actions, such as a purchase, tariff switch, or charging decision, with accuracy and calibration within the relevant groups. Self-reports remain legitimate targets when the research question concerns opinion or experience itself, but matching them cannot establish that people will act on an expressed intention. Our sensor and grade criteria provide external evidence, yet correlations with those records do not demonstrate calibrated behavioural prediction or show that synthetic populations preserve the human gap between intentions and actions \cite{kutzner_consequential_2026}.

Experimental prediction requires the same distinction between the measured outcome and the intended application. Ashokkumar and colleagues show that model predictions can track variation in human treatment effects while exaggerating their magnitude \cite{ashokkumar_large_2026}. Our FSD3587 contrasts likewise reveal substantial errors in the size of responses to different messengers. Our experimental comparisons assess changes in argument ratings. Predicting later behaviour requires separate validation. Copying one synthetic persona into treatment and control conditions would introduce a further claim about that profile's counterfactual response. As the companion framework explains, researchers would first need repeated draws under the same condition to quantify simulation noise, then evidence that differences between conditions correspond to human effects for comparable profiles. Our between-group effect comparisons do not by themselves validate that stronger use.

Representational fairness also requires evidence below the population average. The companion framework asks whether accuracy, response variation, and measurement comparability hold within the groups that a decision affects, and whether researchers document the procedure sufficiently for others to examine the claim \cite{kutzner_consequential_2026}. Cint and Bovitz demonstrate the limits that sparse item-level subgroup samples impose on demographic comparisons. A model that reproduces the average response of renters, for example, could still erase variation in their access to charging facilities and misstate who can respond to a tariff. Researchers should define groups through the decision's relevant constraints, examine their errors and variation, and retain explicit gaps where the human evidence is insufficient. Human--model measurement comparability and comparability between demographic groups require separate tests.

Several limits constrain the reach of these findings. The Twin text comparisons cover the 84 percent of profiles that pass without corrections; bounding missing survey answers cannot establish performance for the excluded people. Their results under other profile conditions do not establish how distilled profiles would perform for them. Some independent human criterion associations are weak, so small absolute correlation errors do not necessarily indicate useful behavioural prediction. Some model sequences stop because an answer falls outside the permitted options, an output is empty or malformed, or a provider rejects the request; these failures limit the questions we can compare. The human sources are public, and we cannot establish whether model training included their questions, answers, or published findings, so the benchmark does not demonstrate generalisation to unseen studies. Models also differ in decoding settings, quantisation, and collection routes; our comparisons describe these evaluated configurations rather than isolating model architecture. Model comparisons therefore apply to the evaluated datasets, endpoints and configurations, subject to the reported uncertainty and coverage.

\section{Conclusion}
A useful synthetic survey must preserve the features of human responses that its intended analysis needs, whether those concern an overall distribution, differences between groups, relationships among traits, or responses to an intervention. The Artificial Societies Benchmark brings eleven tests of those requirements into a common scorecard and shows that a convincing average, consistent answers, or richer profiles can each coexist with consequential failures elsewhere. Researchers can use the framework to choose relevant checks, compare synthetic results with human evidence, and identify the additional data their proposed claims require. When the intended claim concerns what people will do, that evidence must include their behaviour; when it concerns a particular group, the validation must reach that group. The purpose is to make the limits of a synthetic study clear enough that researchers can decide which conclusions it supports.

\bibliographystyle{unsrtnat}
\bibliography{as-benchmark}

@article{tjuatja_llms_2024,
	title = {Do {LLMs} {Exhibit} {Human}-like {Response} {Biases}? {A} {Case} {Study} in {Survey} {Design}},
	volume = {12},
	issn = {2307-387X},
	shorttitle = {Do {LLMs} {Exhibit} {Human}-like {Response} {Biases}?},
	url = {https://direct.mit.edu/tacl/article/doi/10.1162/tacl_a_00685/124261/Do-LLMs-Exhibit-Human-like-Response-Biases-A-Case},
	doi = {10.1162/tacl_a_00685},
	language = {en},
	urldate = {2026-09-21},
	journal = {Transactions of the Association for Computational Linguistics},
	author = {Tjuatja, Lindia and Chen, Valerie and Wu, Tongshuang and Talwalkwar, Ameet and Neubig, Graham},
	month = sep,
	year = {2024},
	pages = {1011--1026},
}

@article{karetnikov_large_2026,
	title = {Large language models as human proxies},
	issn = {2662-8457},
	url = {https://www.nature.com/articles/s43588-026-01060-3},
	doi = {10.1038/s43588-026-01060-3},
	language = {en},
	urldate = {2026-09-22},
	journal = {Nature Computational Science},
	author = {Karetnikov, Nikita and Rahwan, Iyad and Svetinovic, Davor},
	month = sep,
	year = {2026},
}

@incollection{hu_simbench_2026,
	title = {{SimBench}: {Benchmarking} the {Ability} of {Large} {Language} {Models} to {Simulate} {Human} {Behaviors}},
	shorttitle = {{SimBench}},
	url = {https://openreview.net/forum?id=PL51SpN6ZJ},
	urldate = {2026-09-22},
	booktitle = {The {Fourteenth} {International} {Conference} on {Learning} {Representations}},
	author = {Hu, Tiancheng and Baumann, Joachim and Lupo, Lorenzo and Collier, Nigel and Hovy, Dirk and Röttger, Paul},
	year = {2026},
	note = {Version Number: 4},
}

@misc{lukauskas_plausible_2026,
	title = {Plausible but {Not} {Valid}: {A} {Psychometric} {Audit} of {LLMs} as {Synthetic} {Survey} {Respondents}},
	copyright = {arXiv.org perpetual, non-exclusive license},
	shorttitle = {Plausible but {Not} {Valid}},
	url = {https://arxiv.org/abs/2608.14606},
	doi = {10.48550/ARXIV.2608.14606},
	urldate = {2026-09-22},
	publisher = {arXiv},
	author = {Lukauskas, Mantas and Šarkauskaitė, Viktorija},
	month = aug,
	year = {2026},
	note = {Version Number: 1},
}

@inproceedings{shen_reliability_2026,
	address = {San Diego, California, USA},
	title = {The {Reliability} {Illusion} in {Synthetic} {Patients}: {Psychometric} {Misalignment} of {Open}-weight {LLMs} on {PHQ}-9 and {GAD}-7},
	shorttitle = {The {Reliability} {Illusion} in {Synthetic} {Patients}},
	url = {https://aclanthology.org/2026.clpsych-1.7},
	doi = {10.18653/v1/2026.clpsych-1.7},
	language = {en},
	urldate = {2026-09-22},
	booktitle = {Proceedings of the 11th {Workshop} on {Computational} {Linguistics} and {Clinical} {Psychology} ({CLPsych} 2026)},
	publisher = {Association for Computational Linguistics},
	author = {Shen, Qian and Han, Yu},
	year = {2026},
	pages = {88--99},
}

@misc{jia_when_2026,
	title = {When {Can} {Digital} {Personas} {Reliably} {Approximate} {Human} {Survey} {Findings}?},
	copyright = {arXiv.org perpetual, non-exclusive license},
	url = {https://arxiv.org/abs/2605.10659},
	doi = {10.48550/ARXIV.2605.10659},
	urldate = {2026-09-22},
	publisher = {arXiv},
	author = {Jia, Mumin and Chen, Yilin and Sharma, Divya and Diaz-Rodriguez, Jairo},
	year = {2026},
	note = {Version Number: 1},
}

@article{xie_evaluating_2026,
	title = {Evaluating the statistical realism of {LLM}-generated social science data},
	volume = {123},
	issn = {0027-8424, 1091-6490},
	url = {https://pnas.org/doi/10.1073/pnas.2538145123},
	doi = {10.1073/pnas.2538145123},
	language = {en},
	number = {19},
	urldate = {2026-09-22},
	journal = {Proceedings of the National Academy of Sciences},
	author = {Xie, Yueqi and Liang, Lemeng and Li, Shuzhen and Lu, Yifu and Xiao, Zhiwen and Shi, Mengdi and Huang, Junming and Wang, Mengdi and Xie, Yu},
	month = may,
	year = {2026},
	pages = {e2538145123},
}

@inproceedings{wang_sociobench_2025,
	address = {Suzhou, China},
	title = {{SocioBench}: {Modeling} {Human} {Behavior} in {Sociological} {Surveys} with {Large} {Language} {Models}},
	shorttitle = {{SocioBench}},
	url = {https://aclanthology.org/2025.emnlp-main.1335},
	doi = {10.18653/v1/2025.emnlp-main.1335},
	language = {en},
	urldate = {2026-09-22},
	booktitle = {Proceedings of the 2025 {Conference} on {Empirical} {Methods} in {Natural} {Language} {Processing}},
	publisher = {Association for Computational Linguistics},
	author = {Wang, Jia and Zhao, Ziyu and Ni, Tingjuntao and Wei, Zhongyu},
	year = {2025},
	pages = {26268--26300},
}

@article{ashokkumar_large_2026,
	title = {Large language models can predict the results of social science experiments},
	volume = {656},
	issn = {0028-0836, 1476-4687},
	url = {https://www.nature.com/articles/s41586-026-10742-x},
	doi = {10.1038/s41586-026-10742-x},
	language = {en},
	number = {8126},
	urldate = {2026-09-22},
	journal = {Nature},
	author = {Ashokkumar, Ashwini and Hewitt, Luke and Ghezae, Isaias and Willer, Robb},
	month = aug,
	year = {2026},
	pages = {115--122},
}

@article{jalbert_intuitions_2026,
	title = {Intuitions about content moderation are misaligned with effective practices for reducing conspiracy beliefs},
	volume = {4},
	issn = {2731-9121},
	url = {https://www.nature.com/articles/s44271-026-00480-1},
	doi = {10.1038/s44271-026-00480-1},
	language = {en},
	number = {1},
	urldate = {2026-09-22},
	journal = {Communications Psychology},
	author = {Jalbert, Madeline and Wack, Morgan},
	month = may,
	year = {2026},
	pages = {122},
}

@article{orchinik_learning_2024,
	title = {Learning from and about scientists: {Consensus} messaging shapes perceptions of climate change and climate scientists},
	volume = {3},
	copyright = {https://creativecommons.org/licenses/by-nc/4.0/},
	issn = {2752-6542},
	shorttitle = {Learning from and about scientists},
	url = {https://academic.oup.com/pnasnexus/article/doi/10.1093/pnasnexus/pgae485/7857506},
	doi = {10.1093/pnasnexus/pgae485},
	language = {en},
	number = {11},
	urldate = {2026-09-22},
	journal = {PNAS Nexus},
	author = {Orchinik, Reed and Dubey, Rachit and Gershman, Samuel J and Powell, Derek M and Bhui, Rahul},
	editor = {Van Bavel, Jay},
	month = oct,
	year = {2024},
	pages = {pgae485},
}

@misc{park_llm_2026,
	title = {{LLM} {Agents} {Grounded} in {Self}-{Reports} {Enable} {General}-{Purpose} {Simulation} of {Individuals}},
	copyright = {arXiv.org perpetual, non-exclusive license},
	url = {https://arxiv.org/abs/2411.10109},
	doi = {10.48550/ARXIV.2411.10109},
	urldate = {2026-09-22},
	publisher = {arXiv},
	author = {Park, Joon Sung and Zou, Carolyn Q. and Kamphorst, Jonne and Egan, Niles and Shaw, Aaron and Hill, Benjamin Mako and Cai, Carrie and Morris, Meredith Ringel and Liang, Percy and Willer, Robb and Bernstein, Michael S.},
	month = jun,
	year = {2026},
	note = {Version Number: 3},
}

@article{bisbee_synthetic_2024,
	title = {Synthetic {Replacements} for {Human} {Survey} {Data}? {The} {Perils} of {Large} {Language} {Models}},
	volume = {32},
	copyright = {https://creativecommons.org/licenses/by/4.0},
	issn = {1047-1987, 1476-4989},
	shorttitle = {Synthetic {Replacements} for {Human} {Survey} {Data}?},
	url = {https://www.cambridge.org/core/product/identifier/S1047198724000056/type/journal_article},
	doi = {10.1017/pan.2024.5},
	language = {en},
	number = {4},
	urldate = {2026-09-22},
	journal = {Political Analysis},
	author = {Bisbee, James and Clinton, Joshua D. and Dorff, Cassy and Kenkel, Brenton and Larson, Jennifer M.},
	month = oct,
	year = {2024},
	pages = {401--416},
}

@misc{jordan_replication_2026,
	title = {Replication {Data} for: {New} {Evidence} and {Design} {Considerations} for {Repeated} {Measure} {Experiments} in {Survey} {Research}},
	copyright = {, Creative Commons Attribution 4.0 International},
	shorttitle = {Replication {Data} for},
	url = {https://dataverse.harvard.edu/citation?persistentId=doi:10.7910/DVN/SH0F25},
	doi = {10.7910/DVN/SH0F25},
	urldate = {2026-09-22},
	publisher = {Harvard Dataverse},
	author = {Jordan, Diana and Ollerenshaw, Trent and Trexler, Andrew},
	collaborator = {{Harvard Dataverse} and Trexler, Andrew},
	year = {2026},
}

@misc{jessee_replication_2022,
	title = {Replication {Data} for: {Moderator} {Placement} in {Survey} {Experiments}: {Racial} {Resentment} and the “{Welfare}” vs. “{Assistance} to the {Poor}” {Question} {Wording} {Experiment}},
	copyright = {, Creative Commons Zero v1.0 Universal},
	shorttitle = {Replication {Data} for},
	url = {https://dataverse.harvard.edu/citation?persistentId=doi:10.7910/DVN/SZCAZF},
	doi = {10.7910/DVN/SZCAZF},
	urldate = {2026-09-22},
	publisher = {Harvard Dataverse},
	author = {Jessee, Stephen and Albertson, Bethany},
	collaborator = {Jessee, Stephen},
	year = {2022},
}

@misc{cruz_replication_2025,
	title = {Replication {Data} for: {Survey} {Quality} and {Acquiescence} {Bias}: {A} {Cautionary} {Tale}},
	copyright = {, Creative Commons Zero v1.0 Universal},
	shorttitle = {Replication {Data} for},
	url = {https://dataverse.harvard.edu/citation?persistentId=doi:10.7910/DVN/ZF32UF},
	doi = {10.7910/DVN/ZF32UF},
	urldate = {2026-09-22},
	publisher = {Harvard Dataverse},
	author = {Cruz, Andrés and Bouyamourn, Adam and Ornstein, Joseph},
	collaborator = {Cruz, Andrés},
	year = {2025},
}

@misc{condon_selected_2024,
	title = {Selected personality data from the {SAPA}-{Project}: {01Jan2023} to {12Jan2024}},
	copyright = {, Creative Commons Zero v1.0 Universal},
	shorttitle = {Selected personality data from the {SAPA}-{Project}},
	url = {https://dataverse.harvard.edu/citation?persistentId=doi:10.7910/DVN/3BTT82},
	doi = {10.7910/DVN/3BTT82},
	urldate = {2026-09-22},
	publisher = {Harvard Dataverse},
	author = {Condon, David},
	collaborator = {Condon, David},
	year = {2024},
}

@article{vanderweele_promotion_2017,
	title = {On the promotion of human flourishing},
	volume = {114},
	issn = {0027-8424, 1091-6490},
	url = {https://pnas.org/doi/full/10.1073/pnas.1702996114},
	doi = {10.1073/pnas.1702996114},
	language = {en},
	number = {31},
	urldate = {2026-09-22},
	journal = {Proceedings of the National Academy of Sciences},
	author = {VanderWeele, Tyler J.},
	month = aug,
	year = {2017},
	pages = {8148--8156},
}

@article{cronbach_construct_1955,
	title = {Construct validity in psychological tests.},
	volume = {52},
	issn = {1939-1455, 0033-2909},
	url = {https://doi.apa.org/doi/10.1037/h0040957},
	doi = {10.1037/h0040957},
	language = {en},
	number = {4},
	urldate = {2026-09-22},
	journal = {Psychological Bulletin},
	author = {Cronbach, Lee J. and Meehl, Paul E.},
	month = jul,
	year = {1955},
	pages = {281--302},
}

@article{campbell_convergent_1959,
	title = {Convergent and discriminant validation by the multitrait-multimethod matrix.},
	volume = {56},
	issn = {1939-1455, 0033-2909},
	url = {https://doi.apa.org/doi/10.1037/h0046016},
	doi = {10.1037/h0046016},
	language = {en},
	number = {2},
	urldate = {2026-09-22},
	journal = {Psychological Bulletin},
	author = {Campbell, Donald T. and Fiske, Donald W.},
	year = {1959},
	pages = {81--105},
}

@misc{gordon_richer_2026,
	title = {Do {Richer} {Personas} {Improve} {LLM} {Survey} {Simulation}? {A} {Fidelity} {Paradox}},
	copyright = {https://creativecommons.org/licenses/by/4.0/},
	shorttitle = {Do {Richer} {Personas} {Improve} {LLM} {Survey} {Simulation}?},
	url = {https://www.ssrn.com/abstract=7306183},
	doi = {10.2139/ssrn.7306183},
	urldate = {2026-09-22},
	publisher = {SSRN},
	author = {Gordon, Andrew and Petrova, Nora and Burden, John and Bosch, Oriol and Ding, Ning},
	year = {2026},
	note = {SSRN preprint},
}

@misc{kutzner_consequential_2026,
	title = {Consequential {Behaviour} and {Representational} {Fairness} in the {Validation} of {Synthetic} {Research}},
	url = {http://arxiv.org/abs/2609.27690},
	doi = {10.48550/arXiv.2609.27690},
	urldate = {2026-09-24},
	publisher = {arXiv},
	author = {Kutzner, Florian and Kacperski, Celina and Molière, Laura de and Chidichimo, Edoardo and Jung, Min Jun and Wallis, Felix Patrick Sedgwick and He, James Kunling},
	month = sep,
	year = {2026},
	note = {arXiv:2609.27690 [cs.CL]},
}

@article{cohen_coefficient_1960,
	title = {A {Coefficient} of {Agreement} for {Nominal} {Scales}},
	volume = {20},
	copyright = {https://journals.sagepub.com/page/policies/text-and-data-mining-license},
	issn = {0013-1644, 1552-3888},
	url = {https://journals.sagepub.com/doi/10.1177/001316446002000104},
	doi = {10.1177/001316446002000104},
	language = {en},
	number = {1},
	urldate = {2026-09-24},
	journal = {Educational and Psychological Measurement},
	author = {Cohen, Jacob},
	month = apr,
	year = {1960},
	pages = {37--46},
}

@article{sundh_unified_2023,
	title = {A unified explanation of variability and bias in human probability judgments: {How} computational noise explains the mean–variance signature.},
	volume = {152},
	issn = {1939-2222, 0096-3445},
	shorttitle = {A unified explanation of variability and bias in human probability judgments},
	url = {https://doi.apa.org/doi/10.1037/xge0001414},
	doi = {10.1037/xge0001414},
	language = {en},
	number = {10},
	urldate = {2026-09-24},
	journal = {Journal of Experimental Psychology: General},
	author = {Sundh, Joakim and Zhu, Jian-Qiao and Chater, Nick and Sanborn, Adam},
	month = oct,
	year = {2023},
	pages = {2842--2860},
}

@inproceedings{moore_are_2024,
	address = {Miami, Florida, USA},
	title = {Are {Large} {Language} {Models} {Consistent} over {Value}-laden {Questions}?},
	url = {https://aclanthology.org/2024.findings-emnlp.891},
	doi = {10.18653/v1/2024.findings-emnlp.891},
	language = {en},
	urldate = {2026-09-24},
	booktitle = {Findings of the {Association} for {Computational} {Linguistics}: {EMNLP} 2024},
	publisher = {Association for Computational Linguistics},
	author = {Moore, Jared and Deshpande, Tanvi and Yang, Diyi},
	year = {2024},
	pages = {15185--15221},
}

@article{cronbach_coefficient_1951,
	title = {Coefficient {Alpha} and the {Internal} {Structure} of {Tests}},
	volume = {16},
	copyright = {https://www.cambridge.org/core/terms},
	issn = {0033-3123, 1860-0980},
	url = {https://www.cambridge.org/core/product/identifier/S0033312300046494/type/journal_article},
	doi = {10.1007/BF02310555},
	language = {en},
	number = {3},
	urldate = {2026-09-24},
	journal = {Psychometrika},
	author = {Cronbach, Lee J.},
	month = sep,
	year = {1951},
	pages = {297--334},
}

@book{mcdonald_test_1999,
	address = {Mahwah, N.J},
	title = {Test theory: a unified treatment},
	isbn = {978-0-8058-3075-0},
	shorttitle = {Test theory},
	publisher = {L. Erlbaum Associates},
	author = {McDonald, Roderick P.},
	year = {1999},
}

@article{wu_identification_2016,
	title = {Identification of {Confirmatory} {Factor} {Analysis} {Models} of {Different} {Levels} of {Invariance} for {Ordered} {Categorical} {Outcomes}},
	volume = {81},
	copyright = {https://www.cambridge.org/core/terms},
	issn = {0033-3123, 1860-0980},
	url = {https://www.cambridge.org/core/product/identifier/S0033312300015271/type/journal_article},
	doi = {10.1007/s11336-016-9506-0},
	language = {en},
	number = {4},
	urldate = {2026-09-24},
	journal = {Psychometrika},
	author = {Wu, Hao and Estabrook, Ryne},
	month = dec,
	year = {2016},
	pages = {1014--1045},
}

@article{rao_resampling_1988,
	title = {Resampling {Inference} with {Complex} {Survey} {Data}},
	volume = {83},
	issn = {0162-1459, 1537-274X},
	url = {https://www.tandfonline.com/doi/full/10.1080/01621459.1988.10478591},
	doi = {10.1080/01621459.1988.10478591},
	language = {en},
	number = {401},
	urldate = {2026-09-24},
	journal = {Journal of the American Statistical Association},
	author = {Rao, J. N. K. and Wu, C. F. J.},
	month = mar,
	year = {1988},
	pages = {231--241},
}

@misc{toubia_twin-2k-500_2025,
	title = {Twin-{2K}-500: {A} dataset for building digital twins of over 2,000 people based on their answers to over 500 questions},
	shorttitle = {Twin-{2K}-500},
	url = {http://arxiv.org/abs/2505.17479},
	doi = {10.48550/arXiv.2505.17479},
	urldate = {2026-09-24},
	publisher = {arXiv},
	author = {Toubia, Olivier and Gui, George Z. and Peng, Tianyi and Merlau, Daniel J. and Li, Ang and Chen, Haozhe},
	month = may,
	year = {2025},
	note = {arXiv:2505.17479 [cs.CY]},
}

@misc{herne_kaisa_tampere_university_fsd3587_nodate,
	title = {{FSD3587} {Argument} {Quality} {Evaluation} {Study} 2018},
	url = {https://services.fsd.tuni.fi/catalogue/FSD3587},
	doi = {10.60686/T-FSD3587},
	urldate = {2026-09-24},
	publisher = {Finnish Social Science Data Archive (FSD)},
	author = {{Herne, Kaisa (Tampere University)} and {Kestilä-Kekkonen, Elina (Tampere University)} and {Mattinen, Laura (Tampere University)} and {Sipinen, Josefina (Tampere University)}},
}

@misc{european_social_survey_european_research_infrastructure_ess_eric_ess7_2023,
	title = {{ESS7} - integrated file, edition 2.3},
	url = {https://ess.sikt.no/en/datafile/9c96a1b2-b027-43c1-8c74-e883f892d0bb},
	doi = {10.21338/ESS7E02_3},
	urldate = {2026-09-24},
	publisher = {Sikt - Norwegian Agency for Shared Services in Education and Research},
	author = {{European Social Survey European Research Infrastructure (ESS ERIC)}},
	year = {2023},
}

@misc{european_social_survey_european_research_infrastructure_ess_eric_ess7_2016,
	title = {{ESS7} - test variables from {Supplementary} questionnaire, edition 2.1},
	url = {https://ess.sikt.no/en/datafile/fe825d65-9727-49a4-90f8-af3ae0fa57f3},
	doi = {10.21338/ESS7MTMME2_1},
	urldate = {2026-09-24},
	publisher = {Sikt - Norwegian Agency for Shared Services in Education and Research},
	author = {{European Social Survey European Research Infrastructure (ESS ERIC)}},
	year = {2016},
}

@misc{hachen_nethealth_2026,
	title = {{NetHealth} {Project} {Data}},
	copyright = {Creative Commons Attribution 4.0 International},
	url = {https://zenodo.org/doi/10.5281/zenodo.21904040},
	doi = {10.5281/ZENODO.21904040},
	urldate = {2026-09-24},
	publisher = {Zenodo},
	author = {Hachen, David S. and Lizardo, Omar and Striegel, Aaron and Poellabauer, Christian and Milenkovic, Tijana and Chawla, Nitesh V. and Payne, Jessica},
	month = aug,
	year = {2026},
}

@article{bainbridge_evaluating_2022,
	title = {Evaluating the {Big} {Five} as an organizing framework for commonly used psychological trait scales.},
	volume = {122},
	issn = {1939-1315, 0022-3514},
	url = {https://doi.apa.org/doi/10.1037/pspp0000395},
	doi = {10.1037/pspp0000395},
	language = {en},
	number = {4},
	urldate = {2026-09-24},
	journal = {Journal of Personality and Social Psychology},
	author = {Bainbridge, Timothy F. and Ludeke, Steven G. and Smillie, Luke D.},
	month = apr,
	year = {2022},
	pages = {749--777},
}

@misc{davern_gss_2026,
  author = {Davern, Michael and Bautista, Rene and Freese, Jeremy and Herd, Pamela and Morgan, Stephen L.},
  title = {General Social Survey 1972--2024},
  year = {2026},
  publisher = {NORC at the University of Chicago},
  note = {2024 public-use data and codebook, Release 3a},
  url = {https://gss.norc.org/get-the-data},
}

@misc{anes_2020_data,
  author = {{American National Election Studies}},
  title = {{ANES} 2020 Time Series Study Full Release},
  year = {2021},
  note = {Dataset and documentation, February 10, 2022 version},
  url = {https://electionstudies.org/data-center/2020-time-series-study/},
}

@misc{anes_2024_data,
  author = {{American National Election Studies}},
  title = {{ANES} 2024 Time Series Study},
  year = {2026},
  url = {https://electionstudies.org/data-center/2024-time-series-study/},
}

@misc{pew_western_2018,
  author = {{Pew Research Center}},
  title = {Western Europe Survey Dataset},
  year = {2018},
  url = {https://www.pewresearch.org/dataset/western-europe-survey-dataset/},
}

@misc{gfs_waves_data,
  author = {{Global Flourishing Study}},
  title = {Waves 1 and 2 Global Data},
  year = {n.d.},
  url = {https://osf.io/ge2x5/},
  note = {Public annual-wave data, accessed September 2026},
}

@misc{census_htops_2026,
  author = {{U.S. Census Bureau}},
  title = {Household Trends and Outlook Pulse Survey Public Use Files},
  year = {2026},
  note = {March, May, and July 2026 releases},
  url = {https://www.census.gov/programs-surveys/household-pulse-survey/data/datasets.html},
}
\appendix
\section{Metric definitions}
\label{app:metrics}
This appendix collects the formal definitions behind the tests in Section~\ref{sec:benchmark}. All definitions share notation.
\begin{table}[htbp]
\centering\small
\caption{\textbf{Eleven validity tests and their principal diagnostics.}}
\label{tab:tests}
\begin{tabularx}{\textwidth}{@{}llY@{}}
\toprule
Test & Measure & Human-referenced diagnostics\\
\midrule
IV-1 & Repeated questions & Repeated-answer agreement, kappa, and numerical change\\
IV-2 & Linked questions & Constraint violations and their joint patterns\\
IV-3 & Paraphrased questions & Agreement and response-transition distance\\
IV-4 & Perturbed surveys & Response shifts under instrument perturbations\\
CV-1 & Item covariance & Reliability and within-construct item associations\\
CV-2 & Independent criteria & Associations with independently measured criteria\\
CV-3 & Distinct constructs & Between-construct associations and separation\\
EV-1 & Response distributions & Total variation and range-normalised Wasserstein distance\\
EV-2 & Groups and individuals & Conditional distributions, subgroup spread, and person loss\\
EV-3 & Experimental effects & Signed and absolute treatment-effect error\\
EV-4 & Latent structure & Loadings, factor correlations, and ordinal invariance\\
\bottomrule
\end{tabularx}
\end{table}

\paragraph{Notation.} Throughout, $i \in \mathcal{P}$ indexes the $n$ matched respondents of a task, $j \in \mathcal{J}$ indexes administered items, and $k \in \{1,\dots,K_j\}$ indexes the response options of item $j$. The recorded human response is $h_{ij}$ and the synthetic response is $s_{ij}$, with $s^{(r)}_{ij}$ denoting the $r$th of $R$ independent generations. For a categorical item, $p$ and $q$ denote the human and synthetic option-frequency vectors, so $p_k = \frac{1}{n}\sum_i \mathbb{1}[h_{ij}=k]$ and $q_k$ likewise. For numerical items, $W_j$ is the permitted response range, which we use to normalise every loss to $[0,1]$. Human and synthetic versions of a statistic $\theta$ carry superscripts $h$ and $s$, and we report errors as signed differences $\theta^s - \theta^h$ together with their absolute values. A scale $\mathcal{S} \subseteq \mathcal{J}$ contains $m$ items, $T$ denotes the number of scales or factors in an instrument, and $\bar{s}_{i\mathcal{S}} = \frac{1}{m}\sum_{j\in\mathcal{S}} s_{ij}$ is the trait score of person $i$ after we reflect reverse-keyed items. Direct human and synthetic comparisons use matched participants. Factor-transfer diagnostics use the stated human reference cohort.

\subsection{Internal validity}

\paragraph{IV-1 Repeated questions.} For item $j$, let $\mathcal{P}_j$ contain the $n_j$ people with valid answers in both the first and last generation. Item-specific agreement and Cohen's kappa are
\begin{equation}
A_j = \frac{1}{n_j}\sum_{i\in\mathcal{P}_j}
\mathbb{1}\!\left[s^{(1)}_{ij} = s^{(R)}_{ij}\right], \qquad
\kappa_j = \frac{A_j - A_{e,j}}{1 - A_{e,j}}, \qquad
A_{e,j} = \sum_{k=1}^{K_j} q^{(1)}_{jk}\, q^{(R)}_{jk},
\end{equation}
where $q^{(r)}_{jk}$ is the frequency of option $k$ among these people in generation $r$, and $A_{e,j}$ is their expected agreement from the marginal frequencies. Kappa is undefined when $A_{e,j}=1$. Numerical items use $|s^{(1)}_{ij} - s^{(R)}_{ij}|/W_j$, which we average within each person before averaging across people. Human references replace $s^{(1)}$ and $s^{(R)}$ with the two human administrations. Displayed summaries across repeated item pairs average their agreement values with equal pair weights.

\paragraph{IV-2 Linked questions.} Each constraint $c \in \mathcal{C}$ is a linear relation over independently answered items with coefficients $a_{cj}$, constant $b_c$, and tolerance $\tau$. The residual and violation indicator for person $i$ are
\begin{equation}
\begin{aligned}
\varepsilon_{ic} &= \sum_j a_{cj}s_{ij}-b_c,\\
v^s_{ic} &=
\begin{cases}
\mathbb{1}[|\varepsilon_{ic}|>\tau] & \text{for equalities},\\
\mathbb{1}[\varepsilon_{ic}>\tau] & \text{for inequalities}.
\end{cases}
\end{aligned}
\end{equation}
where we define $v^h_{ic}$ identically on $h_{ij}$. The violation rate is $V^s_c = \frac{1}{n}\sum_i v^s_{ic}$ and we report it with $V^s_c - V^h_c$, the mean residual magnitude across matched blocks, the proportion of people with $\max_c v_{ic} = 1$, and the total variation distance between the human and synthetic distributions of the binary pattern $(v_{i1},\dots,v_{i|\mathcal{C}|})$. Magnitude is $|\varepsilon_{ic}|$ for an equality and $\max(\varepsilon_{ic},0)$ for an inequality, including values below the violation tolerance. When a person completes repeated blocks, residual summaries average matched blocks; joint-pattern summaries give equal weight to people and equal weight to their complete blocks. Participant bootstrap draws keep repeated blocks together.

\paragraph{IV-3 Paraphrased questions.} For an original question $o$ and a paraphrase $o'$, let $\pi_{ab}$ be the joint distribution of $(s_{io}, s_{io'})$ over answer pairs $(a,b)$, which we compute with equal weight per person and equal weight per eligible pair within a person. Then
\begin{equation}
A = \sum_{a} \pi_{aa}, \qquad
\kappa = \frac{A - A_e}{1 - A_e}, \qquad
A_e = \sum_{a} \pi_{a\cdot}\,\pi_{\cdot a}, \qquad
D_{\pi} = \frac{1}{2}\sum_{a,b}\left|\pi^s_{ab} - \pi^h_{ab}\right|,
\end{equation}
and the reported quantities are $\kappa^s - \kappa^h$, $A^s - A^h$, and the transition distance $D_{\pi}$. Kappa is undefined when $A_e=1$.

\paragraph{IV-4 Perturbed surveys.} For two administered versions $A$ and $B$ with a scored or range-positioned outcome, the version effect and its error are
\begin{equation}
\Delta^h = \bar{h}_{B} - \bar{h}_{A}, \qquad
\Delta^s = \bar{s}_{B} - \bar{s}_{A}, \qquad
e_{\Delta} = \Delta^s - \Delta^h,
\end{equation}
and for categorical outcomes the mean absolute difference between arm-wise probability-change vectors is $\frac{1}{K}\sum_k \left|(q^{B}_k - q^{A}_k) - (p^{B}_k - p^{A}_k)\right|$. Graded stimuli with ordered levels additionally compare adjacent-level changes, endpoint change, and least-squares slope, together with the fraction of people whose responses decrease at any adjacent step.

\subsection{Construct validity}

\paragraph{CV-1 Item covariance.} For a scale $\mathcal{S}$ of $m$ items after we reflect reverse-keyed items, Cronbach's alpha and McDonald's omega are
\begin{equation}
\alpha = \frac{m}{m-1}\left(1 -
\frac{\sum_{j\in\mathcal{S}} \sigma^2_j}{\sigma^2_{\Sigma}}\right), \qquad
\omega = \frac{\left(\sum_{j\in\mathcal{S}} \lambda_j\right)^2}
{\left(\sum_{j\in\mathcal{S}} \lambda_j\right)^2 +
\sum_{j\in\mathcal{S}}\left(1 - \lambda_j^2\right)},
\end{equation}
where $\sigma^2_j$ is the variance of item $j$, $\sigma^2_\Sigma$ the variance of the scale sum, and $\lambda_j$ the standardised loading from a one-factor fit. The omega expression assumes unit factor variance and uncorrelated residuals. GSS one-factor diagnostics use standardised alpha, $m\bar r/[1+(m-1)\bar r]$, where $\bar r$ is the mean inter-item correlation in the fitted correlation matrix. For a specified family of item pairs $\mathcal{Q}$,
\begin{equation}
\mathrm{MAE}_{\rho} = \frac{1}{|\mathcal{Q}|}\sum_{(j,j')\in\mathcal{Q}}
\left|\rho^s_{jj'} - \rho^h_{jj'}\right|,
\end{equation}
The association estimator follows the source-specific rule. GSS uses polychoric correlations for ordered items, external ordinal and numerical pairs use Spearman correlations, the HTOPS and GFS construct gaps use Pearson correlations, and nominal pairs use Cram\'er's $V$. Pairs require at least 30 matched participants and variation in both human and synthetic responses. HTOPS and GFS summaries retain every specified pair within each diagnostic. We report a summary as unavailable if any required correlation is undefined. These summaries require 90 percent complete model coverage among humans with complete item data and at least 95 percent finite paired bootstrap draws.

The within-trait summary is the mean of $\rho_{jj'}$ over pairs with $j, j'$ in the same scale, which we report as $\bar{\rho}^{\,s}_{\mathrm{within}} - \bar{\rho}^{\,h}_{\mathrm{within}}$.

\paragraph{CV-2 Independent criteria.} Where the source records an independent criterion $y_i$ for the same people,
\begin{equation}
r^{s}_{\mathcal{S}y} =
\mathrm{corr}_{\mathrm{Spearman}}\!\left(\bar{s}_{i\mathcal{S}},\, y_i\right),
\qquad
e_r = r^{s}_{\mathcal{S}y} - r^{h}_{\mathcal{S}y},
\end{equation}
with a paired participant bootstrap resampling both correlations jointly.

\paragraph{CV-3 Distinct constructs.} With $r_{tt'}$ the Spearman correlation between the trait scores of scales $t$ and $t'$,
\begin{equation}
\mathrm{MAE}_{r} = \binom{T}{2}^{-1}\sum_{t<t'}
\left| r^{s}_{tt'} - r^{h}_{tt'} \right|,
\end{equation}
and the between-trait item summary $\bar{\rho}_{\mathrm{between}}$ is the mean of $\rho_{jj'}$ over pairs in different scales. We compare the separation $\bar{\rho}_{\mathrm{within}} - \bar{\rho}_{\mathrm{between}}$ between the synthetic and human samples. Where a fitted measurement model exists, the latent factor correlations $\Phi_{tt'}$ replace $r_{tt'}$.

\subsection{External validity}

\paragraph{EV-1 Response distributions.} For categorical items,
\begin{equation}
\mathrm{TVD}(p,q) = \frac{1}{2}\sum_{k=1}^{K}|p_k - q_k|, \qquad
\mathrm{JSD}(p,q) = \sqrt{\tfrac{1}{2}\,\mathrm{KL}_2(p \,\|\, u) +
\tfrac{1}{2}\,\mathrm{KL}_2(q \,\|\, u)}, \quad u = \tfrac{1}{2}(p+q),
\end{equation}
where $\mathrm{KL}_2$ is the Kullback--Leibler divergence in bits and both quantities lie in $[0,1]$. Supporting diagnostics are the Spearman correlation between $p$ and $q$, $100\sqrt{\frac{1}{K}\sum_k (p_k-q_k)^2}$ in percentage points, and $\mathbb{1}[\arg\max p = \arg\max q]$ where we compare tied winners as complete sets. Numerical items use $W_1(p,q)/W_j$, the Wasserstein distance we divide by the response range, and ordered items report $(\mathrm{SD}^s - \mathrm{SD}^h)/W_j$.

\paragraph{EV-2 Groups and individuals.} For a recorded demographic group $g$, we apply the metrics of EV-1 to the conditional distributions $p_{\cdot\mid g}$ and $q_{\cdot\mid g}$, subject to minimum group sizes. The per-observation loss is
\begin{equation}
\ell_{ij} = \mathbb{1}[s_{ij} \neq h_{ij}] \ \text{for categorical items},
\qquad
\ell_{ij} = \frac{|s_{ij} - h_{ij}|}{W_j} \ \text{for numerical items},
\end{equation}
and the participant-level summary is $\frac{1}{n}\sum_i \frac{1}{|\mathcal{J}_i|}\sum_{j\in\mathcal{J}_i}\ell_{ij}$, where $\mathcal{J}_i$ is the set of items person $i$ answered.

\paragraph{EV-3 Experimental effects.} For a specified contrast between treatment arm $T$ and control arm $C$ with one observation per participant,
\begin{equation}
\hat{\Delta}^h = \bar{h}_{T} - \bar{h}_{C}, \qquad
\hat{\Delta}^s = \bar{s}_{T} - \bar{s}_{C}, \qquad
e_{\Delta} = \hat{\Delta}^s - \hat{\Delta}^h,
\end{equation}
where bars denote arm means of the scored outcome. Intervals resample participants within assignment groups and compute the paired difference of model-minus-human means between arms in each draw.

\paragraph{EV-4 Latent structure.} Let $\Lambda^h$ be the loading matrix of a factor model we fit on the stated human reference cohort, with $\lambda_{jt}$ the loading of item $j$ on factor $t$, and let $\Lambda^s$ and $\Phi^s$ be the loadings and factor correlations from the same structure we fit to the synthetic sample. Then
\begin{equation}
\begin{aligned}
\mathrm{RMSE}_{\lambda} &= \sqrt{\frac{1}{N_\lambda}
\sum_{\lambda^h_{jt}\neq 0} \left(\lambda^s_{jt}-\lambda^h_{jt}\right)^2},\\
\phi_t &= \frac{\langle\lambda^h_{\cdot t},\lambda^s_{\cdot t}\rangle}
{\|\lambda^h_{\cdot t}\|\,\|\lambda^s_{\cdot t}\|},\\
\mathrm{MAE}_{\Phi} &= \binom{T}{2}^{-1}\sum_{t<t'}
\left|\Phi^s_{tt'}-\Phi^h_{tt'}\right|.
\end{aligned}
\end{equation}
where $N_\lambda$ counts the nonzero reference loadings and $\phi_t$ is the Tucker congruence of factor $t$. Congruence requires nonzero loading vectors, and the factor-correlation average requires at least two factors. Factor scores use the reference human standardisation and regression weights. We regress each of the $T$ factor scores on $d$ demographic contrasts, excluding the intercept from the attribute-transfer error $\sqrt{\frac{1}{dT}\sum_{l=1}^{d}\sum_{t=1}^{T}(\beta^s_{lt}-\beta^h_{lt})^2}$. Ordinal invariance fits configural, threshold, loading, and intercept models to the joint human and synthetic data and reports robust nested chi-square differences and changes in fit indices.

\subsection{Common-cohort information comparisons}
Let $\mathcal{P}_D$ be the 430 people whose unedited distilled profile passed the grounding audit. For conditions $a,b$, $\mathcal{J}_i^{ab}$ contains the same human-observed items with valid answers in both conditions. The paired change is
\begin{equation}
 d_i^{ab}=\frac{1}{|\mathcal{J}_i^{ab}|}\sum_{j\in\mathcal{J}_i^{ab}}
 \{\ell(s_{ij}^{b},h_{ij})-\ell(s_{ij}^{a},h_{ij})\},\qquad
 \widehat\delta_{ab}=\frac{1}{|\mathcal{P}_{ab}|}\sum_{i\in\mathcal{P}_{ab}}d_i^{ab},
\end{equation}
where $\mathcal{P}_{ab}\subseteq\mathcal{P}_D$ contains people with a paired loss. Estimation requires at least 30 such people. Each person's difference averages the items models answered validly in both conditions; we report item coverage beside each comparison. These available-item comparisons can use different item sets. We report the audit acceptance fraction separately. Negative values favour condition $b$. Intervals use 500 paired participant resamples. The exclusion check compares mean loss for the 82 excluded people with the included cohort in each available non-distilled condition and compares their recorded-to-verbatim changes. Appendix~\ref{app:twin_sensitivity} bounds the effect of missing answers on comparisons within the accepted cohort.

\subsection{Independent Twin criteria}
For behavioural item $j$ and independent criterion $k$, let $\mathcal{P}_{jk}$ contain people with the human answer, criterion, and valid model answer. With $\rho$ denoting Spearman correlation, the reported error is
\begin{equation}
 e_{jk}=\rho(s_{ij},c_{ik})-\rho(h_{ij},c_{ik}),\qquad i\in\mathcal{P}_{jk}.
\end{equation}
Criteria are the seven-item cognitive-reflection count, five-item numeracy count, 18-item Need for Cognition mean, and five BFI-44 domain means, all from waves 1 to 3. Complete criterion scores use the published item keys and never enter prompts. We score each of 32 wave-four item variants against all eight criteria. A primary error requires 30 matched people, 90 percent complete model coverage, and 95 percent finite bootstrap draws. Constant answers yield an undefined correlation and we record them as collapse. Cell summaries report the mean absolute and signed error over estimable pairs with their count out of 256. These conditional summaries retain the failed-pair count and carry no pooled inferential interval.

\subsection{Missing answers, uncertainty, and interpretation}
For a bounded loss we observe on a fraction $v$ of eligible records, the full-record mean lies in $[v\bar\ell,\,v\bar\ell+(1-v)]$. This bound uses the same record weights as its estimand. Participant bootstrap draws retain all items and conditions and respect assignment groups. Intervals are pointwise. A diagnostic block may contain both estimable and failed endpoints; its presence never certifies the test. Coverage counts study families with at least one estimate for the specified model. Aggregation uses equal item, task, and family weights within one endpoint and condition. We keep different units separate, and report incomplete endpoint support explicitly.

\section{Design details}
\label{app:design}
The L1--L4 information labels below follow Gordon and colleagues' simulation protocol \cite{gordon_richer_2026}.\footnote{These labels describe model inputs and are separate from the companion framework's L0--L3 diagnostic levels, which describe properties of human--synthetic correspondence.}
{\footnotesize
\begin{xltabular}{\textwidth}{@{}p{0.13\textwidth} p{0.11\textwidth} Y p{0.075\textwidth} p{0.09\textwidth} p{0.085\textwidth} p{0.135\textwidth}@{}}
\caption{\textbf{Human sources of the benchmark.} Counts refer to benchmark rosters
after partitioning; separate figures within a cell refer to separate tasks or
cohorts from the same source, which we list in order. Twin entries give the broader task followed by the common information comparison. Information levels follow Gordon
et al.\ \cite{gordon_richer_2026} with the adaptations we describe in
Section~\ref{sec:elicitation}. Sources sharing respondents form one study
family. Tests are those that yield an estimate after the gates of
Section~\ref{sec:scoring}. ESS records category collapse under CV-1 and EV-4.
Cint and Bovitz have limited EV-2 support, with one eligible item-group
combination and none, respectively.}\label{tab:coverage}\\
\toprule
Source & Design & Content & People & Items & Elicitation & Tests \\
\midrule
\endfirsthead
\caption[]{\textbf{Human sources of the benchmark (continued).}}\\
\toprule
Source & Design & Content & People & Items & Elicitation & Tests \\
\midrule
\endhead
\midrule
\multicolumn{7}{r@{}}{\emph{Continued on next page}}\\
\endfoot
\bottomrule
\endlastfoot
GSS 2024 \cite{davern_gss_2026} & National survey & Confidence in twelve institutions; interpersonal
trust, fairness, and helpfulness & 664; 276 & 12; 3 & L1, L2, L3 & IV-1, EV-1,
EV-2, EV-4 \\
Twin-2K-500 \cite{toubia_twin-2k-500_2025} & Online panel, four waves & Behavioural and economic-preference
questions in randomly assigned alternative forms; three written
self-descriptions & 512; 430 & 42; 32 variants, 24; 14 per person & L3, history, L4 & IV-1,
IV-4, CV-1, CV-2, EV-1, EV-2, EV-3 \\
Weather judgments, Exp.~3 \cite{sundh_unified_2023} & Repeated-judgment experiment & Probabilities of
weather events, complements, conjunctions, disjunctions, and three repetitions &
55 & 60 & Unprofiled & IV-1, IV-2, CV-1, EV-1 \\
ValueConsistency \cite{moore_are_2024} & Paraphrase battery & Yes/no value questions with released
paraphrases on 27 topics & 64 & 5 per battery; 89 pairs & Unprofiled & IV-3,
EV-1 \\
Jordan, Ollerenshaw, and Trexler \cite{jordan_replication_2026} & Six randomised experiments & Information, partisan-cue, and wording treatments on policy attitudes & 512 per experiment
& 2 per experiment & L3 & IV-4, EV-1, EV-2, EV-3 \\
Jordan block-order sequence \cite{jordan_replication_2026} & Randomised block order & Forced-short versus
fully randomised ordering and spacing of the same modules & 512 & 23 & L3 &
IV-4, EV-3 \\
Albertson and Jessee \cite{jessee_replication_2022} & Randomised wording & Welfare versus assistance wording
of a spending question & 512 & 2 & L3 & IV-4, EV-1, EV-2, EV-3 \\
Cint and Bovitz panels \cite{cruz_replication_2025} & Randomised polarity & Twenty factual propositions
with reversed polarity and availability of ``Not sure'' & 512 per panel & 20;
80 variants & L3 & IV-4, EV-1, EV-2, EV-3 \\
FSD3587 \cite{herne_kaisa_tampere_university_fsd3587_nodate} & Randomised messenger & Argument ratings under four messenger
identities across four Finnish policy topics & 1,280 & 2 per person; 16
variants & L3 & IV-4, EV-1, EV-2, EV-3 \\
Misaligned Moderation, Exp.~1 \cite{jalbert_intuitions_2026} & Within-person randomised removal & Belief in
social-media posts from which the source removed plausible or implausible evidence, original
images & 234 & 9 posts; 39 variants & L3 & IV-4, EV-3 \\
Orchinik et al. \cite{orchinik_learning_2024} & Graded stimulus & Likelihood judgments under five levels of
scientific consensus & 686 & 1 bundle; 5 levels & L3 & IV-2, IV-4 \\
ANES 2024 \cite{anes_2024_data} & Randomised option order & Attention to politics under forward and
reverse order & 1,630 & 1; 2 variants & L3 & IV-4, EV-1, EV-2 \\
ANES 2020 \cite{anes_2020_data} & Panel with open text; randomised order & Four off-topic group
thermometers under three profile conditions; attention and trust replay &
512; 5,921 & 4; 2 per person & L3, L4; L3 & IV-4, CV-1, EV-1, EV-2, EV-3 \\
Pew Western Europe 2017 \cite{pew_western_2018} & Randomised numeric labels & Ideology under 0--6 or
1--7 scale numbering in France, Germany, and the UK & 1,439; 1,747; 1,460 & 1;
2 variants & Adapted L3 & IV-4, EV-1, EV-2, EV-3 \\
ESS Round 7, Great Britain \cite{european_social_survey_european_research_infrastructure_ess_eric_ess7_2023,european_social_survey_european_research_infrastructure_ess_eric_ess7_2016} & Split-ballot measurement experiment & Political
engagement and immigration under assigned scale, label, and display methods,
with showcards & 512 & 22 per person; 31 variants & L3 & IV-3, IV-4 \\
SAPA \cite{condon_selected_2024} & Personality inventory & Big Five; well-being, sensation-seeking, and
honesty facets & 512 per battery & 20; 9 & L3 & IV-1, IV-2, IV-3, CV-1, CV-2,
CV-3, EV-1, EV-2, EV-4 \\
NetHealth \cite{hachen_nethealth_2026} & Panel with behavioural criteria & BFI-44 across four waves with
course-grade and wearable criteria & 512; 339; 255; 156 & 44 per wave & L3 &
IV-2, IV-3, CV-1, CV-2, CV-3, EV-1, EV-2, EV-4 \\
Bainbridge et al. \cite{bainbridge_evaluating_2022} & Inventory with second inventory & IPIP Big Five with
independently completed BFI-2 criteria, three samples & 194; 319; 309 & 117;
120; 120 & L3 & IV-1, IV-2, IV-3, CV-1, CV-2, CV-3, EV-1, EV-2, EV-4 \\
Global Flourishing Study \cite{gfs_waves_data} & Panel, two annual waves & Twelve flourishing items
in the US, UK, and Australia & 512 per country & 12 & L3 & CV-1, CV-2, CV-3,
EV-1, EV-2 \\
US Census HTOPS 2026 \cite{census_htops_2026} & 2026 cross-sections & Life satisfaction, mental
health, and institutional trust in March, May, and July 2026 & 512 per
cross-section & 15; 15; 19 & L3 & CV-1, CV-3, EV-1, EV-2, EV-4 \\
\end{xltabular}}

\begin{table}[htbp]
\centering\small
\caption{\textbf{Persona information and elicitation conditions.} Levels L1--L4 follow
Gordon et al.\ \cite{gordon_richer_2026}; we declare the remaining rows as extensions.}
\label{tab:levels}
\begin{tabularx}{\textwidth}{@{}p{0.17\textwidth} Y p{0.15\textwidth} p{0.2\textwidth}@{}}
\toprule
Condition & Information we supply to the model & Output & Sources \\
\midrule
L1 aggregate & Survey context, question, and options for the national adult
population & Percentages over options & GSS \\
L2 composition & L1 plus the unweighted demographic composition of the task
roster in five declared age bands & Percentages over options & GSS \\
L3 demographic & One respondent's recorded age, sex, ethnicity, education,
urbanicity, and political affiliation, or the recorded fields a source
provides & One answer & GSS and every profiled task \\
L3 plus history & L3 plus an extractive summary of the same person's earlier
closed-ended answers & One answer & Twin \\
L4 verbatim & L3 plus original free text, being three self-descriptions in
Twin and pre-election likes, dislikes, and problem statements in ANES 2020 &
One answer & Twin, ANES 2020 \\
L4 distilled & L3 plus a seven-field summary of the free text, which one shared preprocessing model extracts as
quotations and we audit against the source,
with and without history & One answer & Twin, 430 people \\
Unprofiled & Question only, no respondent characteristics & One answer or a
JSON array for a battery & Weather, ValueConsistency \\
Sequence & Any of the above, with the same synthetic respondent's earlier
generated answers to the block & One answer per step & ESS, Jordan order,
evidence removal, graded consensus \\
\bottomrule
\end{tabularx}
\end{table}

\begin{table}[htbp]
\centering\small
\caption{\textbf{The nine evaluated systems.} Output caps are tokens for individual answers and
for GSS distribution requests. Local systems answer the hosted request bodies
through vLLM 0.29 on one H100.}
\label{tab:models}
\begin{tabularx}{\textwidth}{@{}l l l Y@{}}
\toprule
Model & Provider, route & Regime & Decoding controls \\
\midrule
\modelid{gpt-5.6-sol} & OpenAI, batch & Thinking disabled & Temperature 1,
reasoning effort none, caps 100/512 \\
\modelid{claude-opus-5} & Anthropic, batch & Thinking disabled & Temperature 1,
thinking off, effort high (provider default), caps 100/512 \\
\modelid{gemini-3.8-flash} & Google, online & Low thinking, separate &
Backend ignores temperature, low thinking, cap 2,048 \\
\modelid{grok-4.3} & xAI, batch & Thinking disabled & Temperature 1, reasoning
off, caps 100/512 \\
\modelid{mistral-small-2603} & Mistral, batch & Thinking disabled & Temperature
1, caps 100/512; both synchronous and Batch image limits refused one ESS continuation \\
\modelid{deepseek-flash} & DeepSeek, online & Thinking disabled & Temperature 1,
thinking off, caps 100/512 \\
\modelid{Qwen3.8-27B} & Local, BF16 & Vendor defaults, thinking
disabled & Temperature 1, top-$k$ 20, top-$p$ 0.95, caps 100/512 \\
\modelid{Gemma-4-31B-it} & Local, BF16 & Vendor defaults, thinking
disabled & Temperature 1, top-$k$ 64, top-$p$ 0.95, caps 100/512 \\
\modelid{Llama-4-Scout-17B-16E} & Local, INT4 weights & Vendor defaults, no
thinking mode & Temperature 1, top-$p$ 0.9, caps 100/512 \\
\bottomrule
\end{tabularx}
\end{table}

\subsection{Coverage by study family}
\label{app:coverage}
\begin{table}[htbp]
\centering\small
\setlength{\tabcolsep}{3pt}
\caption{\textbf{Study families yielding an estimate and study families run.} CV-3 yields four families for OpenAI, Claude, Gemini, and Gemma and five for the other models. Counts otherwise agree across models. A run can yield a diagnostic failure or insufficient support for an estimate.}
\label{tab:counts}
\begin{tabular}{@{}lrrrrrrrrrrr@{}}
\toprule
Test & IV-1&IV-2&IV-3&IV-4&CV-1&CV-2&CV-3&EV-1&EV-2&EV-3&EV-4\\
\midrule
yielding an estimate &5&5&5&12&8&5&4--5&17&14&9&5\\
run &5&5&5&12&9&5&5&17&15&9&6\\
\bottomrule
\end{tabular}
\end{table}

\subsection{Reading the illustrated scorecard}
\label{app:scorecard_guide}
Figure~\ref{fig:scorecard} uses illustrative selections from the completed scorecard to show how each diagnostic works. We use the same source and endpoint for all nine models in a column. These examples do not establish how typical a result is across sources. IV-1 and IV-2 are Bainbridge sample 1 model-minus-human agreement and contradiction rates. IV-3 is ValueConsistency agreement difference. IV-4 is ANES 2024 first-category order-effect error. CV-1 and CV-3 show model-minus-human differences in within-construct correlations and construct separation for the UK Global Flourishing Study (GFS). All nine models yield estimates on the same 493 matched participants and full specified pair families. CV-2 is Twin recorded mean absolute criterion-correlation error over estimable pairs. EV-1 is SAPA Big Five mean item TVD, EV-2 Twin recorded person loss over the full administered target set, and EV-3 FSD3587 mean absolute effect error in rating points. EV-4 reports loading root-mean-square error for SAPA's well-being, sensation-seeking, and honesty facets, using a fixed human reference we fit on participants with complete benchmark battery data. Five models yield estimates; the other four retain unavailable markers. This exploratory comparison does not establish measurement invariance. The figure shows repetition 1. Unavailable selected diagnostics remain visible.

The cells retain endpoint units. A signed value records direction relative to the matched human reference; its interpretation depends on the endpoint. Absolute differences and bounded loss use their own nonnegative scales. For heatmap intensity, we divide the absolute cell value by the largest absolute value the heatmap displays in that column, among the nine models. Zero denotes no discrepancy and one the largest displayed discrepancy; a column of zeros remains zero. This relative shading supports comparisons within each endpoint. It depends on the displayed systems and supplies no cross-test ranking or acceptance threshold. The headings retain domain colours. Unavailable estimates remain unshaded and carry a dash. We draw every model row from the nine-model scorecard. The complete scorecard retains source-level intervals, coverage, and failure statuses beyond these examples.

\section{Source-specific administration and scoring}
\label{app:sources}
This appendix records, for each source, the cohort, the administered instrument with its adaptations, and the scoring rules that Section~\ref{sec:methods} summarises. Tasks follow the gates of Section~\ref{sec:scoring}, with the source-specific rules below. Request counts are planned requests per model unless we state otherwise; they exclude retries and retain unreached sequence steps.

\subsection{Cohorts, eligibility, and reuse}
For most tasks, a practical cap of 512 participants limits inference cost and computation across models, conditions, and repetitions. We selected nine models to represent state-of-the-art proprietary and open-weight systems.

GSS eligibility requires recorded adulthood, a complete mapped profile, a positive source weight, and at least one observed target response; the interpersonal task uses face-to-face and telephone respondents only, giving 276 benchmark people, and the confidence task 664. SAPA includes respondents in recorded age bands from 19 years with adequate recorded English proficiency and complete target batteries. Experimental tasks retain their sources' eligibility and assignment rules. Twin's item catalogue exceeds the number the source administered to any one respondent because the source assigned forms at random. SAPA batteries and Jordan experiments share source populations and are one study family each.

GFS uses the public annual-wave file and selects adults GFS observed in both waves in the United States, United Kingdom, and Australia with recorded online administration, 32,245, 3,617, and 2,580 eligible people, from whom each country's benchmark roster is a hash-ranked sample of 512. GFS collected year-two observations in 2024. Stems and endpoint labels come from the published translation workbooks; we check them across both waves and against the value labels, repeating each group's lead-in for every isolated item. Language is unavailable in the public records, so these are English-majority cohorts without respondent-level language matching.

The project is company-funded scholarly research. We use GFS under its CC BY-NC instrument licence \cite{vanderweele_promotion_2017} and CC BY data licence, ESS under CC BY-NC-SA, FSD3587 and Twin under CC BY 4.0, and NetHealth's BFI-44 under the instrument's documented research-use grant. Bainbridge prompts use public-domain IPIP items, and we compute the BFI-2 criteria locally. We use Llama 4 Scout under the Llama 4 Community Licence. Source data and instruments retain their respective reuse conditions.

\subsection{Prompt reconstruction and recorded personas}
We reconstruct templates from Appendix C of Gordon et al.\ \cite{gordon_richer_2026} with typesetting normalisation, substituting each source's study dates and question content. We take instrument text from released ballots and questionnaires. GSS uses the fielded category labels and the released 2024 confidence stems with their introduction and the fairness preamble. Bainbridge rows drop the Qualtrics export separator. FSD3587 retains the Finnish battery instruction asking about people in general. ANES 2020 transcript labels are the source gate question that the probe follows, and we resolve candidate names and pronouns. We show Jordan drug-recall choices as labels without archive codes. Cint and Bovitz requests carry each panel's recorded collection dates. Recorded and L4 conditions share one response-format instruction. We do not reproduce exact live screens, showcard artwork, and interviewer probes, and a fixed order replaces ValueConsistency's per-person option shuffle, which the source did not record.

GSS harmonisation gives Hispanic origin priority in the ethnicity mapping, retains party leaners as Independents, and maps settlement categories to the protocol's urbanicity categories, with a stricter mapping as a sensitivity analysis. External tasks use the recorded-persona form of L3 in a single user message that keeps the instruction to answer as the person. We do not convert age bands to exact ages and omit unavailable fields. We render SAPA age and proficiency codes as readable descriptions. Weather and ValueConsistency are unprofiled.

\subsection{Twin free text, distillation, and information comparisons}
\label{app:profile_audit}
Twin's three open questions ask respondents who they aspire to be, ought to be, and actually are; all 2,058 source people answered, with a combined median of 117 words. The 512 benchmark people receive six cells, namely demographics alone, demographics with closed-ended history, verbatim descriptions, history with verbatim descriptions, distilled descriptions, and history with distilled descriptions. The demographic and history tasks contain 42 item variants and 24 assigned observations per person. The common information comparison uses the 430 accepted profiles and excludes ten policy self-placement items, because self-descriptions may state political views directly, and retains 32 behavioural item variants with 14 assigned observations per person. The 32 variants are alternative forms of classic judgement problems that the source assigned at random, and we estimate 18 between-person form contrasts (framing, base rate, conjunction, outcome bias, sunk cost, absolute against relative saving, willingness to accept against willingness to pay, Allais, my-side bias, less-is-more, and proportion dominance) with the wording-contrast estimator and read them as EV-3 and IV-4.

Distillation uses one call per person to \modelid{gpt-5.4-2026-03-05} at temperature 0.2 with reasoning disabled, JSON output, and a 1,024-token cap, returning the protocol's seven fields (values summary, moral foundations, decision style, institutional trust, political engagement, key concerns, and consumer orientation). We use each trait field only when the participant explicitly names its subject and it then consists of exact quotations; otherwise it holds the string ``Not stated in the source responses.'' We do not list aspirations and obligations as concerns unless the participant called them one, and the narrative attributes each statement to its question. A second-model audit by \modelid{claude-opus-5}, which we conduct within our study, compares every statement with the source answers, and mechanical exact-quote and target-text checks follow; we exclude profiles that fail without editing, and use the same surviving people and person-item mask for verbatim and distilled comparisons. The shared profile serves every model and repetition. We cap profiles at 16,000 UTF-8 bytes. The profile audit retained 430 of 512 unedited profiles (84.0 percent), yielding 216,720 collected answers across nine models and two repetitions. The audit accepted 453 profiles; 23 required field resets and we excluded them from the unedited cohort. Audit failures include unsupported attributions or quotations; the final cohort also excludes profiles requiring repair.

Within-source information comparisons use paired changes in bounded prediction loss that we average within person, with 500 participant bootstrap draws. Negative changes indicate improved prediction in the second condition. Each comparison requires at least thirty people with a paired difference; item coverage does not determine whether we report it. Each person's difference averages the items models answered validly in both rungs, and we report the lowest item coverage beside every comparison in the ladder file. The minimum across all 270 comparisons is 187 of 216, or 0.866, on a Grok lottery item. All 270 comparisons are estimable, comprising 180 hosted and thirty for each local model. They describe jointly valid answers and can use different item sets; they do not recover outcomes for missing answers. The 82 excluded people form a separate sensitivity check. All twelve of Mistral's individual common-cohort cells also meet the coverage rule.

\subsection{ANES 2020 verbatims and ANES 2024 order}
The February 2022 ANES 2020 release links 8,280 people to eight pre-election redacted text sheets; the three post-election most-important-problem sheets match 7,526. Requiring 100 pre-election words, English administration, a completed post wave, and no identity concern leaves 1,982 eligible people, from whom we hash-select 512 benchmark people. Three conditions compare recorded demographics, demographics with pre-election party and candidate likes and dislikes, and the same text with later problem statements. We retain original strings, and topic patterns we apply to all text exclude potentially disclosed targets in every condition, leaving 1,205 thermometer observations (Jews 432, police 230, transgender people 453, scientists 90). Thermometers take one integer from 0 to 100 with the anchor paragraph that we adapt from person to group. The source measures residence post-election, and the problem statements follow the thermometer battery, so the later conditions are reconstructions. A separate order task replays the attention and trust items for 5,921 benchmark English web respondents (3,021 forward, 2,900 reverse) using pre-election demographics.

ANES 2024 supplies the first substantive question, attention to government and politics, for 2,041 English fresh-web respondents (1,067 forward, 974 reverse); 1,630 form the benchmark. The model receives the stem and labels in the assigned order. We compare reverse-minus-forward category and cumulative probabilities with 500 paired participant bootstraps within arm. Weighted sensitivity uses the full-sample weight with Taylor linearisation over the design variables.

\subsection{Repeatability, retest, and logical coherence (IV-1, IV-2)}
Technical repeatability compares the first and last generation for the same observation, with agreement and Cohen's kappa for categorical items and range-normalised differences for numerical ones. Twin compares earlier human answers with wave-four answers and model answers with the same wave-four records; human persistence is a reference. Bainbridge supplies duplicate-item agreement.

Weather coherence uses 27 linear probability constraints covering complements, conjunctions, disjunctions, inclusion-exclusion, and conditional complements with a one-percentage-point tolerance, reporting violation rates, residual magnitudes, violation-pattern distance, and the share of people with any violation on matched complete blocks. Keyed-opposite pairs in Bainbridge, NetHealth, and SAPA, which we define from item wording, give contradiction rates. Graded consensus contributes the share of people whose stated belief falls as consensus rises and the share of decreasing adjacent steps.

\subsection{Paraphrases and method effects (IV-3, IV-4)}
ValueConsistency supplies 81 attention-passing participants, each answering a five-question battery on one of 27 topics; 64 are in the benchmark. We present the battery in source order and models answer it as a JSON array. A study-specific script applies a fixed list of wording exclusions, set before model results, retaining 89 of 108 original-paraphrase pairs for primary scoring. The exclusions target changes in scope or strength, such as ``everyone'' in place of ``citizens'' or ``necessary'' in place of ``should''. This is a rule-based screen, not an independent human equivalence rating. We display all questions and score all pairs in a sensitivity analysis. We weight agreement, kappa, and transition distance equally over people and pairs within person. Near-paraphrase pairs in Bainbridge, NetHealth, and SAPA, and ESS method pairs, supply further IV-3 estimates.

ESS Round 7 Great Britain administers 13 main and nine assigned supplementary questions in source order, with the current question, its response card where one exists, and the persona's preceding generated answers. Of 2,264 raw records, 2,203 adults with known regional wording remain (form A 1,150, form B 1,053); we hash-select 512 benchmark people (276 form A, 236 form B), twenty of them with ``the UK'' wording. Fifteen card images preserve source orientation, numbers, and labels. We compare paired response transitions, rank associations, and relative category position, stratifying form contrasts by assigned form; 11,264 planned requests per model. Unreached questions stay in coverage denominators. We retain the original request specification and count unreached steps as missing when a provider limit stops a sequence. We report completion and stopping reasons in the \hyperref[app:sequence_completion]{sequence-completion summary}. The common parser accepts displayed codes with matching labels in either order, punctuation, and wrapper markup. The parsing and retry policy below applies to every model.

FSD3587 supplies 1,600 Finnish adults with 3,200 argument-validity ratings on a 0 to 10 scale; 1,280 are in the benchmark. Twenty-four pairwise source-identity contrasts hold argument and topic fixed. We use source Finnish text within the recorded-persona wrapper.

Jordan's block-order sequence preserves 4,287 participants and 81,456 questionnaire pages with randomised block order, repeated items, recall questions, and distractor batteries. We select 512 benchmark participants (256 per ordering procedure), one generation per model, 9,709 planned requests per model, and scores six post-response category contrasts and six pre-post change contrasts with 500 paired draws within procedure.

\subsection{Experimental contrasts (EV-3, IV-4)}
For each contrast we estimate the difference in mean scored response between assigned treatment and control for human and synthetic samples on the same participants and compare it through signed and absolute error, with participant bootstrap within arm. Ordered outcomes use fixed score mappings; the nominal affirmative-action outcome uses favour probability. Acquiescence analyses reverse negative propositions to a common endorsement direction, scoring ``Not sure'' 0.5, and separate contrasts measure uncertainty-response incidence where that option is available.

Both acquiescence sources use split forms and have limited within-item demographic support for EV-2. Bovitz has 512 participants, but each item has only 5 to 41 human respondents within an age or education group. No group meets the required 50 human and matched respondents. Cint yields an EV-2 estimate because just one of 400 item-group combinations reaches this floor. The family-level count therefore records an estimate with narrow support. Both sources are underpowered for subgroup fidelity and provide the acquiescence contrasts for which their authors designed them under EV-3 and IV-4.

Jordan modules retain the source's post-only restriction. Albertson and Jessee uses the question-first subgroup and infers assignment from the administered outcome, as in the source. We reproduce stimuli in fresh requests without the preceding survey blocks.

Misaligned Moderation Experiment 1 reproduces 300 participants under the authors' exclusions; 234 are in the benchmark. Each person sees three topics that the source counterbalanced across all-posts, plausible-removal, and implausible-removal conditions with the nine original post images that we encode losslessly. Models generate seven interest and three truth judgements sequentially. Truth uses the six source positions and belief is seven minus the position. Contrasts compare each removal condition with all posts and the two removal conditions with each other, within person and we average them over the counterbalance, with 500 paired draws within allocation cell; 2,340 planned requests per model.

Orchinik et al.\ supplies a five-slider question at 50, 75, 90, 97, and 99 percent hypothetical consensus. The 847 control-group people of the 2024 journal sample yield 686 benchmark people. All five rows appear together in ascending order and one request returns five integers on the 0 to 100 scale. We compare adjacent changes, the 50-to-99 change, and the within-person slope, with curve error and the decrease shares as secondary summaries; 686 planned requests per model.

Pew's 2017 Western Europe survey assigned 5,840 respondents in France, Germany, and the United Kingdom to ideology scales that the survey numbered 0 to 6 or 1 to 7; the benchmark holds 1,439, 1,747, and 1,460. Profiles contain age, gender, education, and country. We shift displayed 1 to 7 answers to common positions 0 to 6. We compare arm distributions, normalised mean position, midpoint and endpoint mass, and the seven-position effect vector, the half-$L_1$ norm of the difference in effect vectors giving the distributional effect error. Humans could volunteer nonresponse where models choose a substantive integer, so effect comparisons condition on substantive human answers.

\subsection{Distributions, relationships, and subgroups (EV-1, EV-2, CV-1)}
Categorical marginals use total variation distance, GSS adding Jensen-Shannon distance, frequency RMSE, frequency correlation, and plurality agreement. Numerical marginals use range-normalised Wasserstein distance and signed dispersion difference. Individual loss is disagreement for categories and range-normalised absolute error for numbers, which we average within person first. GSS human sampling references use 2,000 pairs of independent bootstrap samples at each item's observed size.

Item relationships use polychoric correlations for ordered GSS items, Cram\'er's $V$ for nominal items, and Spearman correlations for external ordinal or numerical pairs, on the first trial only, requiring 30 matched people with variation on both sides. Convergent-minus-discriminant Pearson correlation gaps for HTOPS and GFS compare within-construct with between-construct item correlations. Each summary retains its full specified pair family in the human data, model data, and every bootstrap draw. Undefined required correlations make the summary unavailable. We require 90 percent complete model coverage among humans with complete item data and at least 95 percent finite bootstrap draws. Subgroups use recorded demographic categories with at least 50 matched people, GSS additionally requiring 100 respondents and an effective size of 50, and compare conditional distributions through total variation and normalised Wasserstein distance together with the spread of group means.

Population coverage is specific to the endpoint. A family we count as yielding an estimate can have eligible support for only one item-group combination. Interpret subgroup results alongside their human and matched counts, uncertainty, and excluded combinations. These eligibility rules govern estimation; a group lacking support retains an unresolved validity claim. The present analysis does not provide every diagnostic within every intersection or a general within-group calibration procedure.

\subsection{Traits, criteria, and factor structure (CV-1 to CV-3, EV-4)}
For the SAPA facets item ``Tell the truth'', Gemma requested a phrase to rate in 478 of 512 responses, and DeepSeek declined 39 times. We retain the item as we administered it. Gemma's facets cells have no estimate; its Big Five cells do.

SAPA uses published keys for the twenty-item Big Five and three three-item facets (well-being, sensation seeking, and honesty) on a six-point scale; NetHealth the BFI-44 and Bainbridge the IPIP Big Five on five-point scales. We score reverse-keyed items as lower plus upper minus response, and normalise trait-score error by item count times response span. We compare scale means, Cronbach's alpha, scale correlations, and within-trait against between-trait item associations with 500 paired draws, retaining negative alpha and reranking ties within draws.

NetHealth links four BFI-44 administrations (719, 415, 317, and 192 respondents in waves 1, 2, 6, and 8) to course grades and wearable measures. Sensor criteria average days 1 to 30 after the survey with at least 14 valid days and 80 percent device compliance; grade criteria average all available letter-graded courses with equal course weights and require at least three courses. Bainbridge's three samples (253, 398, and 388 adults) supply IPIP targets and independently completed BFI-2 domain means as criteria, with 117, 120, and 120 keyed items. GFS supplies year-two prediction against the same person's observed year-one answer and the published six domain, ten-item, and twelve-item composites. We compare human and model trait-criterion Spearman correlations on the same people with paired bootstrap intervals; we report constant model traits as failures.

Criterion provenance determines the scope of interpretation. Independently completed personality scales remain self-report criteria; wearable records and recorded grades provide different forms of external evidence. A trait-criterion correlation measures an association and supplies no direct estimate of person-level predictive calibration. Likewise, performance on hypothetical choices, incentivised decisions, and field behaviour should retain separate interpretations. The current scorecard does not estimate a common intention--behaviour gap or establish transfer from laboratory choices to field outcomes. Human--model ordinal invariance addresses measurement comparability between those samples; subgroup invariance requires its own supported fit.

Factor structures use the published simple structures with correlated factors. For the SAPA, NetHealth, Bainbridge, and HTOPS exploratory comparisons, we fit a fixed human reference on people with complete benchmark battery data. We compare each synthetic sample with this reference on loadings, congruence, factor correlations, and omega. Demographic-slope comparisons use matched people and the same fixed reference transformation. Each output records the reference sample size and the model-specific complete-case count. For DeepSeek on SAPA facets, these counts are 512 and 473, respectively. These ULS comparisons use a continuous approximation and require at least 100 complete matched people and 90 percent complete model coverage among humans with complete battery data. ULS fits use Pearson correlations and require optimiser convergence, off-diagonal correlation RMSE at most 0.10, and no loading above 0.994. Loadings range from 0.001 to 0.995, and optimisation permits at most 1,000 function evaluations. We report poor human reference fits with their indices and label partial or poor model fits. Demographic slopes can remain available when a loading fit fails, so family-level EV-4 coverage can reflect only part of the diagnostic set.

Ordinal measurement invariance requires at least 200 people and fits configural, threshold, loading, and intercept models to joint human and synthetic data with WLSMV estimation and theta parameterisation in \texttt{lavaan}, Wu-Estabrook identification \cite{wu_identification_2016}, robust nested comparisons, and 500 paired bootstrap refits of the separate human and model factor models with seed 9142026. These refits quantify uncertainty in correlation and reliability diagnostics. Nested joint invariance comparisons use robust asymptotic inference; ESS category omissions prevent that joint sequence here. Reference fit requires scaled CFI at least 0.90, RMSEA and SRMR at most 0.08; we report a reference outside these bounds with its indices and flag it. We report a synthetic sample whose responses do not span the scale's categories as a category collapse. We fit the ESS structure reference on 397 development participants over the nine 0 to 10 items. All nine models omit response categories. Humans use all eleven immigration response categories, whereas OpenAI uses only 6 to 10; on the white-immigrants item, model responses concentrate on 0 to 4. The 478 humans with complete battery data also omit category 9 on that item. This omission persists in each model's matched sample and limits the full-category joint fit. The scorer reports category collapse when model categories are absent and leaves categories unmerged. This failure to reproduce scale use contributes evidence to CV-1 and EV-4.

\subsection{2026 cross-sections (HTOPS)}
The March, May, and July 2026 HTOPS files contain 12,521, 12,636, and 12,755 respondents from whom HTOPS collected responses 13 to 30 March, 29 April to 18 May, and 15 July to 3 August 2026. Each cross-section yields 512 benchmark people. Personas contain recorded unallocated demographics and broad geography; we omit imputed fields and the age boundary codes. We administer fifteen items in March and May and nineteen in July individually with their shared preambles, life satisfaction on the displayed 0 to 10 scale. Person-weight sensitivity uses the full weight and all 80 successive-difference replicates with the $4/80$ variance coefficient and 95 percent normal intervals.

\subsection{Baselines and missing outputs}
\label{app:baselines}
GSS baselines are development-sample empirical distributions, demographic multinomial regression, and nearest-donor answer vectors; overlapping development evaluations use five-fold participant cross-fitting. As in the statistical-control comparisons of Lukauskas and \v{S}arkauskait\.{e} \cite{lukauskas_plausible_2026}, we distinguish independent item sampling from sampling that approximates dependence. We fit the Gaussian rank copula on at least 30 complete development people with empirical marginals, midrank Gaussian scores, and fixed 0.05 identity shrinkage, and it draws one joint vector per held-out person with a fixed seed; the independent-marginal control draws from the same marginals. We report invalid and missing model outputs beside conditional scores, with worst-case bounds for bounded losses. Weighted GSS sensitivities use a Rao-Wu rescaled bootstrap over source strata and clusters \cite{rao_resampling_1988}.

\subsection{Analysis scope, parsing, and collection}
We kept internal frozen analysis plans but did not formally preregister the benchmark.

We specified the supplementary item-pair, construct-separation, structural, and Twin form and criterion diagnostics after model answers became available and interpret them descriptively, with pointwise uncertainty intervals. Consensus monotonicity contributes to IV-2, ESS method pairs to IV-3, GFS annual prediction to CV-2, and technical repeats to IV-1. ESS method pairs change several administration features, so we interpret them as method comparisons.

We developed additional parsing rules from observed response formats and applied the final rules to every model. The parser maps responses to a unique permitted option using response text, option labels, and specified normalisations, including documented source-spelling aliases; it does not use the human target answer. Permitted normalisations cover wrappers, markers, case, and punctuation, a prefix that identifies exactly one option, or a whole option label within a longer reply with a negation guard and no mention of a second option. The source-spelling aliases map the exact corrected forms of ``miillion'' and ``probablity'' to the source's option labels; they do not permit general typo correction or paraphrases. Block-order parsing accepts option labels or integer codes within matrix objects or arrays, brace sets, and string-valued ``null''. Consensus parsing accepts one prose preamble before an otherwise valid response array. For example, the parser accepts ``My answer is strongly agree'' when it identifies only that option, but rejects ``Agree'' when both ``Agree a little'' and ``Agree strongly'' remain possible. These rules identify a permitted label; they do not guarantee that every longer reply expresses an unambiguous intention. We preserve raw outputs, record each normalisation, and retain invalid and missing responses in coverage counts.

The batch rescore under the additional option-mapping rules recovered 2,179 previously invalid answers: 2,119 for Mistral, 31 for Grok, 20 for Claude, and nine for DeepSeek, with none for the other five models. These counts concern that batch rescore only; they exclude other formatting rules and sequence recovery.

The shared re-ask policy permits one identical request for an empty output or an invalid GSS aggregate distribution. It applies to 37 Qwen ESS outputs, four Qwen GSS distributions, and one Mistral GSS distribution. Six Twin requests whose connections dropped before any response also received one repeat, five Gemini and one DeepSeek. Hosted collection uses the batch and synchronous routes in Table~\ref{tab:models}, with identical request bodies on both routes. The Mistral ESS request that exceeded the image limit failed through both routes. Remaining failures retain their coverage consequences; we do not estimate the effect of collection route on sampling variability.

\subsection{Local model results}
Llama 4 Scout, Qwen3.8-27B, and Gemma 4 31B received the same question content, profile content, and response-format instructions through model-specific vLLM configurations. Each model returned all 24,080 distilled answers for the common cohort of 430 people. After the shared parsing rules, Scout has one truncated answer, Qwen three invalid options, and Gemma no invalid answers. All 30 common-cohort ladder comparisons per local model are estimable. All 36 local Twin criterion cells have an estimated summary, and we retain failed item-criterion pairs.

\section{Detailed results and completion}
\label{app:detailed_results}
The scorecard contains 7,128 task, model, condition, repetition, and test cells, of which 3,186 contain diagnostics. At the family level, the only measured combinations without an estimate are ESS under CV-1 and EV-4 and Bovitz acquiescence under EV-2. Individual tasks and endpoints within other families can still lack estimates.

\paragraph{Response processes and measurement.} Bainbridge sample 1 human duplicate agreement is approximately 0.73, with model agreement from 0.54 to 0.99; seven models exceed the human reference. Human keyed-opposite contradictions are about 0.08, against 0.005 to 0.24 across models. ValueConsistency agreement is about 0.85 for humans, 0.90 for Scout, and 1.00 for the other eight models. Under increasing consensus, approximately 48 percent of human sequences contain a decrease in belief, while all nine models have zero. DeepSeek, Grok, Mistral, and Qwen yield supported March HTOPS construct-separation estimates, with gaps from $-0.09$ to $0.07$ against about 0.20 for humans. Constant responses make the complete comparison unavailable for OpenAI, Claude, and Gemma. Gemini and Scout have too few finite bootstrap estimates.

Across 108 Twin model-condition-repetition cells, mean absolute criterion correlation error is approximately 0.08 to 0.10, based on 168 to 256 estimable pairs out of 256. Signed error ranges from approximately $-0.03$ to $+0.01$. The largest human association is about 0.28; OpenAI's recorded-condition Allais-choice association with numeracy is about 0.11 against this human value. These associations use eight independent criteria and 32 behavioural item variants, with the scorecard retaining failed or constant-response pairs.

\paragraph{Distributions and effects.} First-generation SAPA Big Five mean item total variation distance ranges from 0.38 to 0.70 across models. Rank-copula and independent-marginal controls reach 0.047 and 0.054, with between-trait correlation errors of 0.040 and 0.170. Mean absolute FSD3587 messenger-effect errors are 0.57 for OpenAI, 1.10 for Claude, 0.50 for Gemini, 0.23 for Grok, 0.45 for Mistral, 0.42 for DeepSeek, 0.58 for Qwen, 0.66 for Gemma, and 0.42 for Scout, on a 0 to 10 scale. Human effects average 0.52 points in absolute size. The complete scorecard retains every contrast and its uncertainty.

\subsection{Paired information comparisons}
\label{app:ladder_results}
Table~\ref{tab:ladder_changes} gives selected paired changes. The common cohort retains 430 of 512 profiles, or 84.0 percent, yielding 216,720 collected distilled answers across nine models, with 24,080 per model. The 82 excluded people's mean loss differs from the included cohort by approximately $-0.01$ to $0.03$ across available conditions, with intervals often including zero. Acceptance and prediction accuracy measure different properties of the simulation procedure.

\begin{table}[htbp]
\centering\small
\caption{\textbf{Change in Twin prediction loss from demographic profiles.} Each cell gives
repetitions 1 and 2, in that order. Negative values favour the added information.
All paired comparisons are estimable; intervals appear
in Figure~\ref{fig:ladder} and the scorecard.}
\label{tab:ladder_changes}
\begin{tabularx}{\textwidth}{@{}lYYY@{}}
\toprule
Model & Original text & Distilled text & History and distilled text\\
\midrule
GPT-5.6 Sol & $-0.014$, $-0.010$ & $-0.019$, $-0.020$ & $-0.020$, $-0.021$\\
Claude Opus 5 & $-0.014$, $-0.010$ & $-0.008$, $-0.006$ & $-0.027$, $-0.028$\\
Gemini 3.8 Flash & $-0.030$, $-0.026$ & $-0.023$, $-0.019$ & $-0.047$, $-0.041$\\
Grok 4.3 & $-0.001$, $+0.000$ & $-0.006$, $+0.003$ & $-0.009$, $-0.008$\\
Mistral Small 2603 & $+0.006$, $+0.009$ & $+0.009$, $+0.009$ & $+0.015$, $+0.012$\\
DeepSeek Flash & $-0.011$, $+0.003$ & $+0.002$, $+0.006$ & $+0.002$, $+0.010$\\
Qwen3.8-27B & $-0.005$, $+0.003$ & $+0.001$, $-0.010$ & $-0.009$, $+0.002$\\
Gemma 4 31B & $-0.009$, $-0.010$ & $-0.003$, $-0.005$ & $-0.008$, $-0.011$\\
Llama 4 Scout & $+0.013$, $+0.018$ & $-0.012$, $-0.005$ & $+0.007$, $+0.011$\\
\bottomrule
\end{tabularx}
\end{table}

\paragraph{Sequence completion.}\label{app:sequence_completion} ESS completion is 512 of 512 for OpenAI, Claude, Gemini, Grok, DeepSeek, Scout, and Gemma, 501 for Mistral with ten off-card stops and one request exceeding the provider's eight-image limit, and 504 for Qwen with seven empty-output stops and one off-card answer. Block-order completion is 512 for OpenAI, Claude, Grok, Mistral, and Gemma, 511 for Gemini, 508 for DeepSeek and Qwen, and 499 for Scout. Gemini has one page without a returned candidate; DeepSeek has three off-card answers and one malformed JSON response. Scout has 9,603 valid pages out of 9,709, or 0.989 coverage, with malformed matrix rows, duplicate selections, and an invalid option among the stops. All twelve block-order contrasts are estimable for every model.

\paragraph{Generation.} Llama 4 Scout, Qwen3.8-27B, and Gemma 4 31B received the same question content, profile content, and response-format instructions as the hosted models through model-specific vLLM configurations, using the shared distilled profiles. GSS, Twin, and SAPA Big Five include two generations; repetitions never count as additional people.

\subsection{Twin missing-answer sensitivity analysis}
\label{app:twin_sensitivity}
The primary paired comparison averages errors only where both conditions provide a valid answer. To assess its sensitivity to unanswered questions, we preserve the same 430-person cohort and every human-observed target in the six-condition comparison. For an observed model answer in condition $c$, let $L^c_{ij}=U^c_{ij}=\ell^c_{ij}$; for an invalid or missing answer, let $L^c_{ij}=0$ and $U^c_{ij}=1$. With $\mathcal{H}_i$ the full set of human-observed target items for person $i$, conservative bounds on the full-target mean difference between conditions $a$ and $b$ are
\begin{equation}
 D^-_{ab}=\frac{1}{430}\sum_i\frac{1}{|\mathcal{H}_i|}\sum_{j\in\mathcal{H}_i}(L^b_{ij}-U^a_{ij}),\qquad
 D^+_{ab}=\frac{1}{430}\sum_i\frac{1}{|\mathcal{H}_i|}\sum_{j\in\mathcal{H}_i}(U^b_{ij}-L^a_{ij}).
\end{equation}
We hold every observed error fixed, including an error from one condition when the other lacks an answer. Each person retains equal weight. These bounds allow any missing loss from zero to one; they do not impute an answer or assume that missing answers resemble observed answers. They address missing responses within the accepted cohort, not the 82 people whose profiles do not qualify.

For sampling uncertainty, we resample the 430 people jointly across both bound endpoints 500 times with seed 9132026. The 2.5th percentile of the resampled lower endpoint and the 97.5th percentile of the resampled upper endpoint form a descriptive bootstrap envelope. This is a pointwise sensitivity envelope, not a simultaneous confidence band across comparisons. All 270 comparisons retain 430 contributing people; the minimum item coverage is 187 of 216, or 86.6 percent.

Across all 270 comparisons, the sample bounds exclude zero for 256, while the bootstrap envelopes exclude zero for 102. The primary intervals exclude zero for 117; fifteen of those comparisons have sensitivity envelopes that include zero. These counts summarise dependent descriptive comparisons rather than a family of confirmatory tests. Table~\ref{tab:twin_sensitivity} reports the envelopes for the three contrasts in Figure~\ref{fig:ladder}. The OpenAI and Gemini distilled-profile improvements, and the Claude and Gemini history-plus-distilled improvements, retain negative envelopes in both repetitions. Mistral's history-plus-distilled sample bounds are $[0.0083,0.0191]$ and $[0.0067,0.0175]$, but the corresponding envelopes include zero. Scout's distilled-versus-original-text envelopes remain negative in both repetitions. Missingness therefore leaves the principal improvement examples intact while weakening several directional conclusions, including Mistral's worsening.

\begin{table}[htbp!]
\centering\small
\caption{\textbf{Twin missing-answer sensitivity: bootstrap envelopes for changes in full-target prediction loss relative to demographics.} Each cell gives repetitions 1 and 2, respectively. An entirely negative range supports improvement throughout the envelope; a range crossing zero leaves direction uncertain. These envelopes include both missing-answer bounds and participant resampling, unlike the primary intervals in Figure~\ref{fig:ladder}.}
\label{tab:twin_sensitivity}
\begin{tabularx}{\textwidth}{@{}lYYY@{}}
\toprule
Model & Original text & Distilled text & History and distilled text\\
\midrule
GPT-5.6 Sol & \shortstack[l]{$[-0.0237,-0.0058]$\\$[-0.0211,-0.0008]$} & \shortstack[l]{$[-0.0289,-0.0090]$\\$[-0.0303,-0.0107]$} & \shortstack[l]{$[-0.0303,-0.0104]$\\$[-0.0315,-0.0095]$}\\[3pt]
Claude Opus 5 & \shortstack[l]{$[-0.0225,-0.0051]$\\$[-0.0180,-0.0013]$} & \shortstack[l]{$[-0.0163,+0.0025]$\\$[-0.0150,+0.0042]$} & \shortstack[l]{$[-0.0369,-0.0165]$\\$[-0.0378,-0.0176]$}\\[3pt]
Gemini 3.8 Flash & \shortstack[l]{$[-0.0396,-0.0209]$\\$[-0.0352,-0.0158]$} & \shortstack[l]{$[-0.0311,-0.0141]$\\$[-0.0280,-0.0101]$} & \shortstack[l]{$[-0.0567,-0.0375]$\\$[-0.0502,-0.0305]$}\\[3pt]
Grok 4.3 & \shortstack[l]{$[-0.0133,+0.0149]$\\$[-0.0149,+0.0138]$} & \shortstack[l]{$[-0.0189,+0.0080]$\\$[-0.0116,+0.0163]$} & \shortstack[l]{$[-0.0232,+0.0059]$\\$[-0.0218,+0.0068]$}\\[3pt]
Mistral Small 2603 & \shortstack[l]{$[-0.0101,+0.0240]$\\$[-0.0075,+0.0262]$} & \shortstack[l]{$[-0.0077,+0.0249]$\\$[-0.0053,+0.0276]$} & \shortstack[l]{$[-0.0048,+0.0319]$\\$[-0.0050,+0.0298]$}\\[3pt]
DeepSeek Flash & \shortstack[l]{$[-0.0221,+0.0004]$\\$[-0.0087,+0.0144]$} & \shortstack[l]{$[-0.0082,+0.0124]$\\$[-0.0051,+0.0166]$} & \shortstack[l]{$[-0.0089,+0.0140]$\\$[-0.0028,+0.0246]$}\\[3pt]
Qwen3.8-27B & \shortstack[l]{$[-0.0187,+0.0088]$\\$[-0.0100,+0.0165]$} & \shortstack[l]{$[-0.0116,+0.0144]$\\$[-0.0227,+0.0009]$} & \shortstack[l]{$[-0.0212,+0.0028]$\\$[-0.0127,+0.0144]$}\\[3pt]
Gemma 4 31B & \shortstack[l]{$[-0.0180,+0.0001]$\\$[-0.0196,-0.0013]$} & \shortstack[l]{$[-0.0131,+0.0065]$\\$[-0.0143,+0.0039]$} & \shortstack[l]{$[-0.0208,+0.0039]$\\$[-0.0240,+0.0006]$}\\[3pt]
Llama 4 Scout & \shortstack[l]{$[+0.0026,+0.0242]$\\$[+0.0076,+0.0286]$} & \shortstack[l]{$[-0.0216,-0.0029]$\\$[-0.0140,+0.0031]$} & \shortstack[l]{$[-0.0051,+0.0184]$\\$[-0.0014,+0.0216]$}\\[3pt]
\bottomrule
\end{tabularx}
\end{table}

\section{Declarations}
\label{app:declarations}
\paragraph{Source acknowledgements.} Pew Research Center bears no responsibility for the analyses or interpretations of the data presented here. The opinions expressed herein, including any implications for policy, are those of the author and not of Pew Research Center. ANES, the original data collectors, and their funders bear no responsibility for our use of the data or the interpretations and inferences we draw.

\paragraph{Funding.} Artificial Societies funded this work.

\paragraph{Author contributions.} E.C. and J.K.H. conceptualised the study. E.C. conducted the research and analyses and drafted the manuscript. F.P.S.W. and J.K.H. jointly supervised the project. M.J.J., F.P.S.W., and J.K.H. reviewed the manuscript.

\paragraph{Ethics.} We analysed existing human datasets and collected new responses from language models. We did not recruit new human participants or obtain institutional ethics approval for this study. Appendix~\ref{app:sources} describes the human data sources and their access conditions.

\paragraph{Data and code availability.} The repository at \url{https://github.com/artificial-societies/artificial-societies-benchmark} provides the analysis code, task specifications, prompt templates, model settings and benchmark results. Readers can obtain the original human datasets through the providers described in Appendix~\ref{app:sources}, subject to their access conditions.

\end{document}